\documentclass[10pt,journal,compsoc]{IEEEtran}
\usepackage[pagebackref=true,breaklinks=true,letterpaper=true,colorlinks,citecolor=blue,linkcolor=blue,bookmarks=false]{hyperref}
\usepackage{amsmath, amssymb, amsfonts, amsthm, fouriernc, mathtools}
\usepackage{tikz}
\tikzstyle{arrow} = [thick,->,>=stealth]
\usepackage{xifthen}
\DeclareMathOperator*{\argmin}{arg\,min}

\usepackage{graphicx}
\graphicspath{ {./pics/} {./eps/}}
\usepackage{epsfig}
\usepackage{epstopdf}
\newcommand{\norm}[1]{\left\lVert#1\right\rVert}
\def\layersep{2.5cm}
\include{macro}
\usepackage{array, booktabs, makecell}
\ifCLASSOPTIONcompsoc
\usepackage[nocompress]{cite}
\else
\usepackage{cite}
\fi

\ifCLASSINFOpdf
\else
\fi
\begin{document}
\title{An overview of deep learning based methods for unsupervised and semi-supervised  
	anomaly detection in videos}

\author{B Ravi Kiran,
Dilip Mathew Thomas,
Ranjith Parakkal
\IEEEcompsocitemizethanks{
\IEEEcompsocthanksitem Preprint of an article under review at \href{http://www.mdpi.com/journal/jimaging}{MDPI Journal of Imaging} 
\IEEEcompsocthanksitem Bangalore Ravi Kiran is with Uncanny Vision Solutions, Bangalore and CRIStAL Lab, UMR 9189, Universit\'e Charles de Gaulle, Lille 3. Email: \href{mailto:beedotkiran@gmail.com}{beedotkiran@gmail.com}
\IEEEcompsocthanksitem Dilip Mathew Thomas and Ranjith Parakkal, are with Uncanny Vision Solutions, Bangalore. Email: \href{mailto:dilip@uncannyvision.com}{dilip@uncannyvision.com}, \href{mailto:ranjith@uncannyvision.com}{ranjith@uncannyvision.com}
}
}

\markboth{}%
{Shell \MakeLowercase{\textit{et al.}}: Bare Demo of IEEEtran.cls for Computer Society Journals}
%



\IEEEtitleabstractindextext{%
\begin{abstract}
Videos represent the primary source of information for surveillance applications 
and are available in large amounts but in most cases contain little or no annotation
for supervised learning. This article reviews the state-of-the-art deep learning based methods  
for video anomaly detection and categorizes them based on the type of model and criteria of detection.  
We also perform simple studies to understand the different approaches and provide the criteria 
of evaluation for spatio-temporal anomaly detection.
\end{abstract}

}

\maketitle

\IEEEdisplaynontitleabstractindextext

%
\IEEEpeerreviewmaketitle


\label{sec:introduction}
%
%
%





\section{Introduction} 
Unsupervised representation learning has become an important domain with the advent of deep generative
models which include the variational  autoencoder (VAE) \cite{kingmaVAE2014ICLR} , generative adversarial
networks (GANs) \cite{goodfellow2014generative}, Long Short Term memory networks 
(LSTMs) \cite{hochreiter1997long} , and others. Anomaly detection is a well-known sub-domain of
unsupervised learning in the machine learning and data mining community. Anomaly
detection for images and videos are challenging due to their high dimensional
structure of the images, combined with the non-local temporal variations across frames.

We focus on reviewing firstly, deep convolution architectures for
feature or representation learnt ``end-to-end'' and secondly, predictive and generative models
specifically for the task of \textit{video anomaly detection}. 
Anomaly detection is an unsupervised learning task
where the goal is to identify abnormal patterns or motions in data
that are by definition infrequent or rare events. Furthermore, anomalies
are rarely annotated and labeled data rarely available to train
a deep convolutional network to separate normal class from the anomalous class.
This is a fairly complex task since the class of normal points includes frequently
occurring objects and regular foreground movements while the anomalous class
include various types of rare events and unseen objects that could be summarized as a consistent class.
Long streams of videos containing no anomalies are made available using which
one is required to build a representation for a moving window over the video stream 
that estimates the normal behavior class while detecting anomalous movements and appearance, such as 
unusual objects in the scene.

Given a set of training samples containing no anomalies, 
the goal of anomaly detection is to design or learn a feature representation,  
that captures ``normal'' motion and spatial appearance patterns. 
Any deviations from this normal can be identified by measuring the 
approximation error either geometrically in a vector space
or the posterior probability of a given model which fits training sample representation vectors 
or by modeling the conditional probability of future samples given their past values and 
measuring the prediction error of test samples by training a predictive model, thus accounting for
temporal structure in videos.  

\subsection{Anomaly detection}  
Anomaly detection is an unsupervised pattern recognition task that can be defined 
under different statistical models. In this study we will explore 
models that perform linear approximations by PCA, non-linear approximation by various
types of autoencoders and finally deep generative models. 

Intuitively, a complex system under the action of various transformations
is observed, the normal behavior is described through a few samples and a statistical model is built
using the said normal behavior samples that is capable of generalizing well on unseen samples. 
The normal class distribution $\mathcal{D}$ is estimated using the training samples $\mathbf{x}_i \in X_\text{train}$, 
by building a representation $f_\theta : X_\text{train} \to \mathbb{R}$ which minimizes model prediction loss
\begin{equation}
\theta^{\ast} = \argmin_{\theta} \sum_{\mathbf{x}_i \in X_\text{train}} L_\mathcal{D}(\theta; \mathbf{x}_i) = 
\argmin_{\theta} \sum_{\mathbf{x}_i \in X_\text{train}} \|f_\theta(\mathbf{x}_i)-\mathbf{x}_i\|^2
\end{equation}

error over all the training samples, over all $i$, is evaluated.
Now the deviation of the test samples $\mathbf{x}_j \in X_\text{test}$ under this representation $f_{\theta^\ast}$ 
is evaluated as the anomaly score, $a(\mathbf{x}_j) = \|f_{\theta^\ast}(\mathbf{x}_j)-\mathbf{x}_j\|_2$ 
is used as a measure of deviation. 
For said models, the anomalous points are samples that are poorly approximated by the estimated model $f_{\theta^\ast}$. 
Detection is achieved by evaluating a threshold on the anomaly score $a_j > T_\text{thresh}$. 
The threshold is a parameter of the detection algorithm and the variation of the threshold w.r.t
detection performance is discussed under the Area under ROC section.
For probabilistic models, anomalous points can be defined as samples that lie in low density or concentration
regions of the domain of an input training distribution $P(\mathbf{x}|\theta)$. 

Representation learning automates feature extraction for
video data for tasks such as action recognition, action similarity, 
scene classification,  object recognition, semantic video segmentation 
\cite{fayyaz2016stfcn},  human pose estimation, human behavior recognition and 
various other tasks. Unsupervised learning tasks in video include anomaly detection
\cite{UCSD_mahadevananomaly2010}, \cite{UCSD_mahadevananomaly2014_PAMI},  
unsupervised representation learning \cite{srivastava2015unsupervisedLSTM}, 
generative models for video \cite{vondrick2016generating}, and video prediction \cite{mathieu2015deep}. 

\begin{figure*}[ht]
	\centering
	Training Samples \hspace{2cm} Test Samples \hspace{2cm} Anomaly Mask \\
	\includegraphics[width=0.85\textwidth]{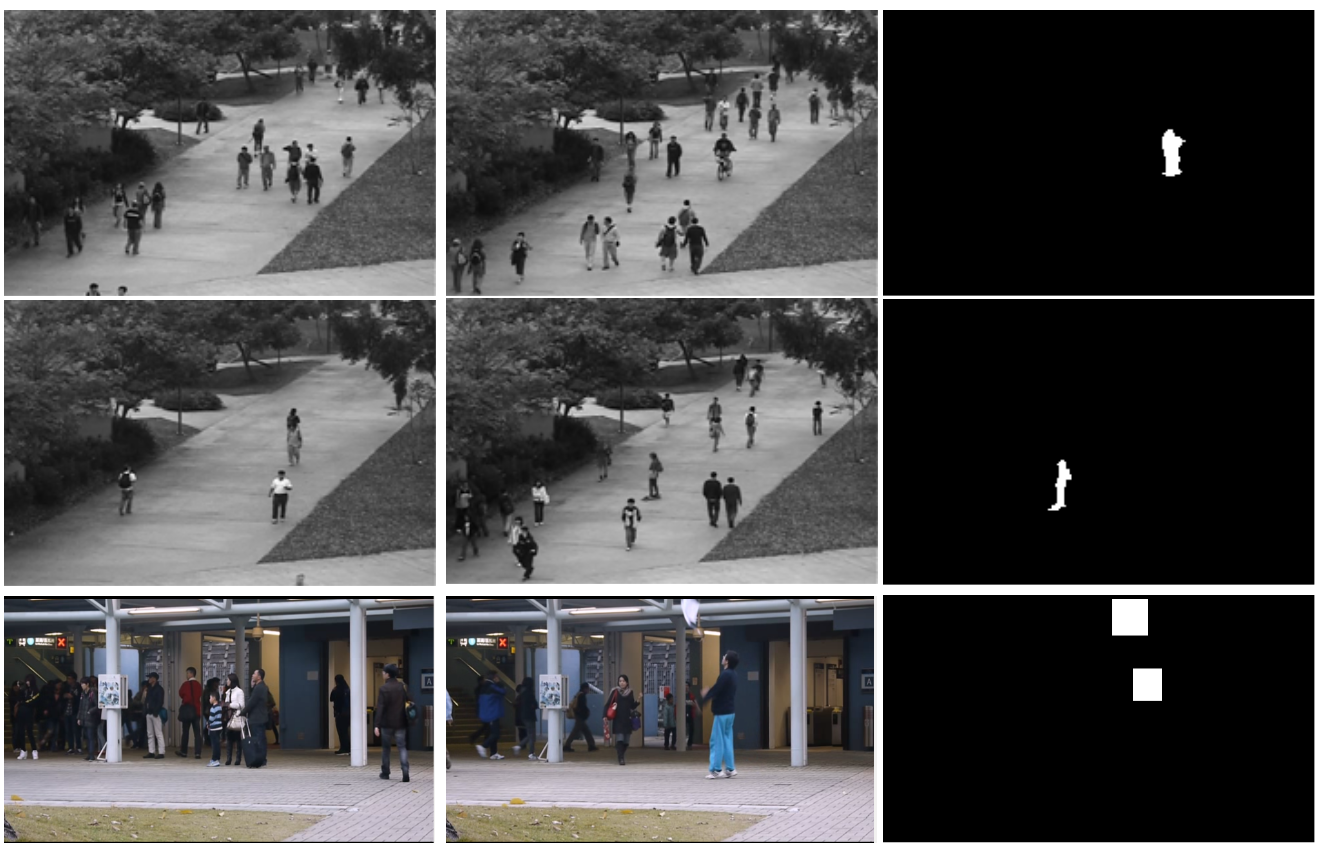}
	\caption{UCSD dataset (top two rows) : Unlike normal training video streams, anomalies consist of 
		a person on a bicycle and a skating board, ground truth detection shown in anomaly mask. Avenue dataset (bottom row) : 
		Unlike normal training video streams, anomalies consist of a person throwing papers. Other examples include 
		walking with an abnormal object (bicycle), a person running (strange action), and a person walking in the wrong direction.}
	\label{fig:examples}
\end{figure*}

\subsection{Datasets} 
We now define the video anomaly detection problem setup. 
The videos considered come from a surveillance camera 
where the background remains static, while the foreground 
constitutes of moving objects such as pedestrians, traffic and so on. The anomalous events 
are the change in appearance and motion patterns that deviate from the normal 
patterns observed in the training set. We see  a few examples 
demonstrated in figure \ref{fig:examples} : 

Here we list the frequently evaluated datasets, though this 
is not exhaustive. The \href{http://www.svcl.ucsd.edu/projects/anomaly/dataset.html}{UCSD dataset} 
\cite{UCSD_mahadevananomaly2010} consists of pedestrian videos where anomalous time 
instances correspond to the appearance of objects like a cyclist, a wheelchair, and a car in 
the scene that is usually populated with pedestrians walking along the
roads. People walking in unusual locations are also considered anomalous. 
In \href{http://www.cse.cuhk.edu.hk/leojia/projects/detectabnormal/dataset.html}{CUHK Avenue Dataset} 
\cite{abnormal2013lu} anomalies correspond to strange actions such as 
a person throwing papers or bag, moving in unusual directions, and appearance of unusual objects 
like bags and bicycle. In the Subway entry and exit 
\href{http://vision.eecs.yorku.ca/research/anomalous-behaviour-data/}{datasets} 
people moving in the wrong direction, loitering and so on are considered as anomalies.  
UMN dataset \cite{umndatasetanomaly} consists of videos showing unusual crowd activity, 
and is a particular case of the video anomaly detection problem. 
The Train dataset \cite{zaharescu2010anomalous} contains moving people in a train. 
The anomalous events are mainly due to unusual movements of people in the train.
And finally the Queen Mary University of London U-turn dataset \cite{benezeth2009abnormal}
contains normal traffic with anomalous events such as jaywalking and movement of a fire engine.
More recently, a controlled environment based LV dataset has been introduced by \cite{leyva2017lv}, 
with challenging examples for the task of online video anomaly detection.

\section{Representation learning for Video Anomaly Detection(VAD)}
Videos are high dimensional signals with both spatial-structure, 
as well as local temporal variations.
An important problem of anomaly detection in videos is to learn a
representation of input sample space $f_\theta : \mathcal{X} \to \mathbb{R}^d$, 
to $d$-dimensional vectors. The idea of feature learning is to automate the process of finding 
a good representation of the input space, that takes into account 
important prior information about the problem \cite{shalev2014understanding}. 
This follows from the No-Free-Lunch-Theorem which states that no universal 
learner exists for every training distribution $\mathcal{D}$.
Following work already established for video anomaly detection, the task concretely consists in 
detecting deviations from models of static background, normal crowd appearance and 
motion from optical flow, change in trajectory and other priors.
Representation learning consists of building a parameterized model 
$f_\theta: \mathcal{X} \to \mathcal{Z} \to \mathcal{X}$, and in this study 
we focus on representations that reconstruct the input, while the latent space 
$\mathcal{Z}$ is constrained to be invariant to changes in the input, such as change in luminance, 
translations of objects in the scene that don't deviate normal movement patterns, and others. This provides
a way to introduce prior information to reconstruct normal samples.

\subsection{Taxonomy}
The goal of this survey is to provide a compact review of the state of
the art in video anomaly detection based on unsupervised and semi-supervised deep 
learning architectures. The survey characterizes the underlying video representation or model as
one of the following :
\begin{enumerate}
	\item \textbf{Representation learning for reconstruction} : Methods such as Principal 
	component analysis (PCA), Autoencoders (AEs) are used to represent the different linear 
	and non-linear transformations to the appearance (image) or motion (flow), 
	that model the normal behavior in surveillance videos. Anomalies represent any 
	deviations that are poorly reconstructed.
	\item \textbf{Predictive modeling} : Video frames are viewed as temporal 
	patterns or time series, and the goal is to model the conditional distribution 
	$P(\mathbf{x}_t|(\mathbf{x}_{t-1}, \mathbf{x}_{t-2},...,\mathbf{x}_{t-p}))$. 
	In contrast to reconstruction, where the goal is to learn a generative model 
	that can successfully reconstruct frames of a video, the goal here is to predict 
	the current frame or its encoded representation using the past
	frames. Examples include autoregressive models and convolutional 
	Long-Short-Term-Memory models. 
	\item \textbf{Generative models} : Variational Autoencoders (VAE), Generative 
	Adversarial Networks (GAN) and Adversarially trained AutoEncoders (AAE), 
	are used for the purpose of modeling the likelihood of normal video samples  
	in an end-to-end deep learning framework.  
\end{enumerate}

An important common aspect in all these models is the problem of representation learning, which refers to the feature extraction
or transformation of input training data for the task of anomaly detection.
We shall also remark the other secondary feature transformations performed in each of these different models and their purposes.

\subsection{Context of the review} 
A short review on the subject of video anomaly detection is provided here  
\cite{chong2015VidAnomalyReview}. To the best of our knowledge, there has not been a systematic study of 
deep architectures for video anomaly detection, which is characterized by abnormal appearance and 
motion features, that occur rarely. We cite below the other domains which do not fall under this study.  

\begin{itemize}
 	\item A detailed review of abnormal human behavior and crowd motion analysis
is provided in \cite{popoola2012video_review} and \cite{li2015crowdedSurvey}. This includes
deep architectures such as Social-LSTM \cite{alahi2016socialLSTM} based on the social 
force model \cite{helbing1995social} where the goal is to predict pedestrian motion
taking into account the movement of neighboring pedestrians.
	\item Action recognition is an important domain in computer vision which requires supervised learning
of efficient representations of appearance and motion for the purpose of classification \cite{kantorov2014efficient}.
Convolutional networks were employed to classify various actions in video quite early \cite{baccouche2011sequential}. 
Recent work involves fusing feature maps evaluated on different frames (over time) of video 
\cite{karpathy2014large} yielding state of the art results. Finally, convolutional networks using 3-D filters (C3D) 
have become a recent base-line for action recognition \cite{tran2015learning_C3D}.
	\item Unsupervised representation learning is a well-established domain and the readers
	are directed towards a complete review of the topic in \cite{bengio2013representation}, 
	as well as the deep learning book \cite{Goodfellow-et-al-2016}.
 \end{itemize} 

In this review, we shall mainly focus on the taxonomy provided and restrict our review to 
deep convolutional networks and deep generative models that enable end-to-end spatio-temporal 
representation learning for the task of anomaly detection in videos.
We also aim to provide an understanding of what aspects of detection do these different models 
target. Apart from the taxonomy being addressed in this study, there have been many other approaches.
One could cite work on anomaly detection based on K-Nearest Neighbors \cite{saligrama2012video}, 
unsupervised clustering \cite{abuolaim2017essence_clustering}, and object speed and size \cite{basharat2008learning}.

We briefly review the set of hand-engineered features used for the task of video anomaly detection,
though our focus still remains deep learning based architectures.
Mixture of dynamic textures (MDT) is a generative mixture model defined for each spatio-temporal windows or cubes
of the raw training video	\cite{UCSD_mahadevananomaly2010}, \cite{UCSD_mahadevananomaly2014_PAMI}. It models
appearance and motion features and thus detects both spatial and temporal anomalies.
Histogram of oriented optical flow and oriented gradients \cite{BinZhaoFeiFei2011CVPR} is a baseline
used in anomaly detection and crowd analysis, \cite{Locmonitors2008}, \cite{HOOF_2013}, \cite{HOFOM_2015}.
Tracklets are representation of movements in videos and have been applied to abnormal crowd motion analysis 
\cite{mousavi2015analyzing}, \cite{mousavi2015abnormality}.
More recently, there has been work on developing optical flow acceleration features
for motion description \cite{OpticalAcc_CVPRW2017}.

\begin{figure}[htbp]
	\centering
	\includegraphics[width=0.5\textwidth]{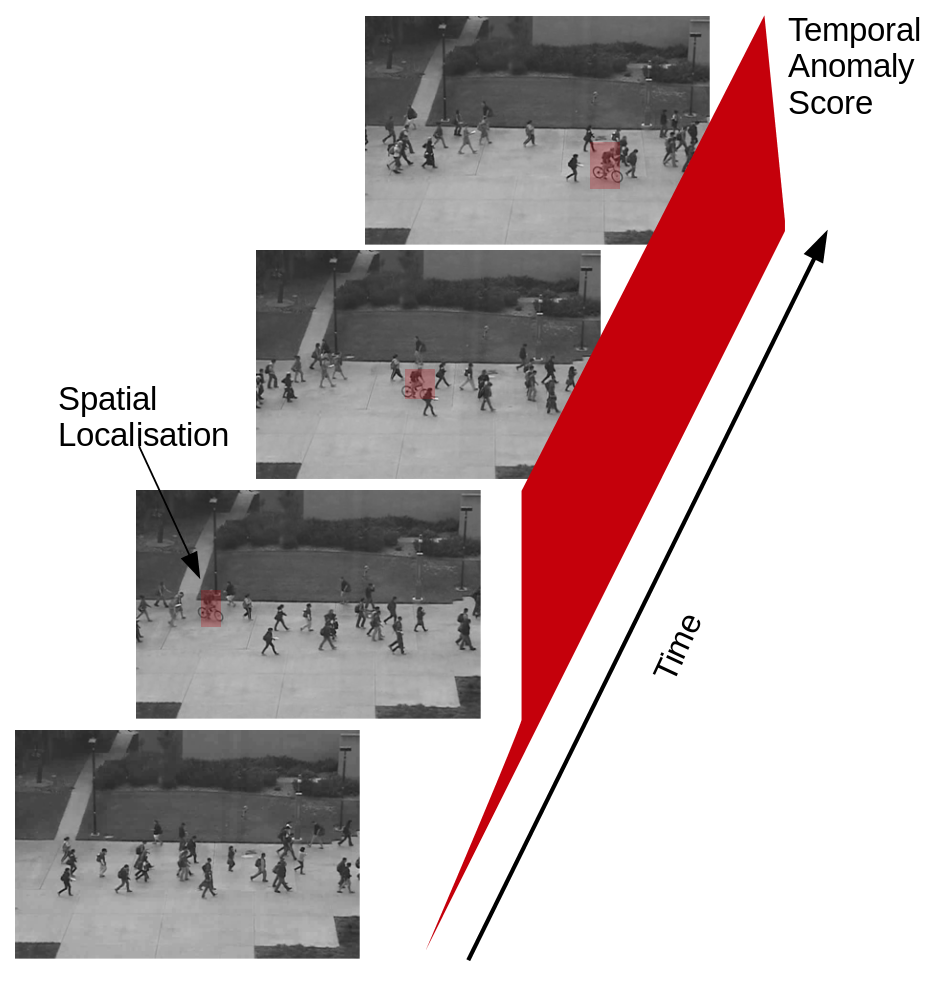}
	\caption{Visualizing anomalous regions and temporal anomaly score.}
	\label{fig:videoAnomalyVisual}
\end{figure}

\noindent \textbf{Problem setup} :
Given a training sequence of images from a video, $X_\text{train} \in \mathbb{R}^{N_\text{train} \times r \times c}$, 
which contains only ``normal motion patterns'' and no anomalies, and given a test sequence 
$X_\text{test} \in \mathbb{R}^{N_\text{test} \times r \times c}$, which is susceptible to contain anomalies, 
the task consists in associating each frame with an anomaly score
for the temporal variation, as well as a spatial score to localize the anomaly in space. This is demonstrated
in figure \ref{fig:videoAnomalyVisual}.

The anomaly detection task is usually considered unsupervised when there is no
direct information or labels available about the positive rare class. However the
samples in the study with no anomalies are available, and thus is a semi-supervised learning problem. 

For $Z = \{\mathbf{x}_i, y_i\}, i \in [1, N]$ we have samples only with $y_i=0$.
The goal thus of anomaly detection is two-fold : first, find the representation of 
the input feature $f_\theta(\mathbf{x}_i)$,  for example, using convolutional neural networks (CNNs), 
and then the decision rule $s(f_\theta(\mathbf{x}_i)) \in \{0,1\}$ that detects 
anomalies, whose detection rate can be parameterized as per the application.

\section{Reconstruction Models}
We begin with an input training video $X_\text{train} \in \mathbb{R}^{N \times d}$, with $N$ frames and $d=r \times c$
pixels per frame, which represents the dimensionality of each vector. In this section, we shall 
focus on reducing the expected reconstruction error by different methods. We shall describe
the Principal Component Analysis (PCA), Convolutional AutoEncoder (ConvAE), and Contractive AutoEncoders (CtractAE), 
and their setup for dimensionality reduction and reconstruction.

\subsection{Principal Component Analysis}
PCA finds the directions of maximal variance in the training data. In the case of videos, 
we are aiming to model the spatial correlation between pixel values which are components of 
the vector representing a frame at a particular time instant.

With input training matrix $X$, which has zero mean, we are looking for a
set of orthogonal projections that whiten/de-correlate the features in the training set : 

\begin{equation}
	\min_{W^T W = I} \|X - (XW)W^T\|_F^2 = \|X-\hat{X} \|_F^2
\end{equation}
where, $W \in \mathbb{R}^{d \times k}$, with the constraint $W^T W=I$ representing an orthonormal
reconstruction of the input $X$. The projection $XW$ is a vector in a lower dimensional subspace, 
with fewer components than the vectors from $X$. This reduction in dimensionality is used to 
capture the anomalous behavior, as samples that are not well reconstructed. The anomaly score is given by the 
Mahalanobis distance between the input and the reconstruction, or the variance scaled reconstruction error : 

\begin{equation}
	A = (X-\hat{X})\Sigma^{-1}(X-\hat{X})^T 
\end{equation}

We associate each frame with continual optical flow magnitude, and learn atomic motion patterns
with standard PCA on the training set, and evaluate reconstruction error on the test optical flow magnitude. 
This serves a baseline for our study. A refined version was implemented and evaluated in
\cite{kim2009observe} which evaluated the atomic movement patterns using a probabilistic PCA \cite{tipping1999mixtures}, 
over rectangular regions over the image domain.

Optical flow estimation is a costly step of this algorithm, and there has been large progress
in the improving its evaluating speed. Authors \cite{wulff2015efficient}, propose to trade of
accuracy of for fast approximation of optical flow using PCA to interpolate flow fields.

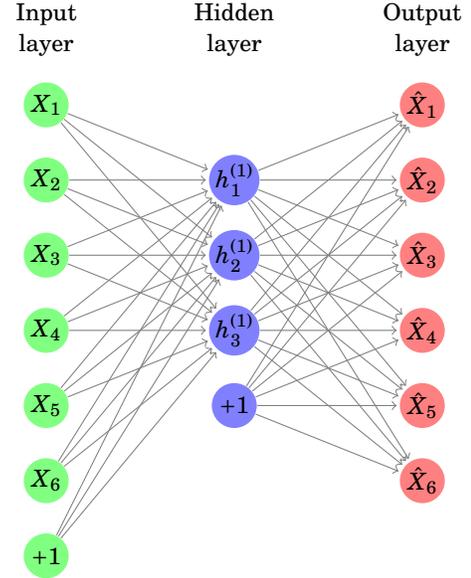
\begin{figure}[htbp]
	\centering
	\begin{tikzpicture}[shorten >=1pt,->,draw=black!50, node distance=\layersep]
	    \tikzstyle{every pin edge}=[<-,shorten <=1pt]
	    \tikzstyle{neuron}=[circle,fill=black!25,minimum size=17pt,inner sep=0pt]
	    \tikzstyle{input neuron}=[neuron, fill=green!50];
	    \tikzstyle{output neuron}=[neuron, fill=red!50];
	    \tikzstyle{hidden neuron}=[neuron, fill=blue!50];
	    \tikzstyle{annot} = [text width=4em, text centered]

	    \foreach \name / \y in {1,...,6}
	        \node[input neuron] (I-\name) at (0,-\y) {$X_\y$};
	    \node[input neuron] (I-7) at (0,-7) {$+1$};
	    
	    \foreach \name / \y in {1,...,6}
	    	\node[output neuron] (O-\name) at (5,-\y) {$\hat{X}_\y$};

	    \foreach \name / \y in {1,...,3}
	        \path[yshift=-1cm]
	        node[hidden neuron] (H-\name) at (\layersep,-\y cm) {$h_\y^{(1)}$};
	    \path[yshift=-1cm]
	    	node[hidden neuron] (H-4) at (\layersep,-4 cm) {$+1$};

	    \foreach \source in {1,...,7}
	        \foreach \dest in {1,...,3}
	            \path (I-\source) edge (H-\dest);

	    \foreach \source in {1,...,4}
	        \foreach \dest in {1,...,6}
	            \path (H-\source) edge (O-\dest);

	    \node[annot,above of=H-1, node distance=2cm] (hl) {Hidden layer};
	    \node[annot,left of=hl] {Input layer};
	    \node[annot,right of=hl] {Output layer};
	\end{tikzpicture}
	\caption{Autoencoder with single hidden layer.}
	\label{fig:autoencoder}
\end{figure}

\subsection{Autoencoders} 
An Autoencoder is a neural network trained by back-propagation and provides an alternative to PCA
to perform dimensionality reduction by reducing the reconstruction error on the training set,
shown in figure \ref{fig:autoencoder}. It takes an input $\mathbf{x} \in \mathbb{R}^d$
and maps it to the latent space representation $\mathbf{z} \in \mathbb{R}^k$, by
a deterministic application, $\mathbf{z} = \sigma(W \mathbf{x} + b)$. 

Unlike the PCA the autoencoder (AE) performs a non-linear point-wise transform of the input 
$\sigma : \mathbb{R} \to \mathbb{R}$, which is required to be a differentiable function. It is 
usually a rectified linear unit (ReLU) ($\sigma(x)=\max(0,x)$) or Sigmoid ($\sigma(x) = (1+e^{-x})^{-1}$). 
Thus we can write a similar reconstruction of the input matrix given by : 

\begin{equation}
 	\min_{U, V} \|X-\sigma(XU)V\|^2_F
 \end{equation} 

The low-dimensional representation is given by $\sigma(XU^\ast)$, where $U^\ast$ represents the optimal
linear encoding that minimizes the reconstruction loss above.
There are multiple ways of regularizing the parameters $U, V$.  
One of the constraints is the average value of the activations in the hidden layer, this enforces sparsity. 

\subsection{Convolutional AutoEncoders (CAEs)}

Autoencoders in their original form do view the input as a signal decomposed as the sum of other signals. 
Convolutional AutoEncoders (CAEs) \cite{masci2011stacked}, 
makes this decomposition explicit by weighting the result of the convolution operator.
For a single channel input $\mathbf{x}$ (for example a gray-scale image),
the latent representation of the $k$th filter would be :

\begin{equation}
	h^k = \sigma(\mathbf{x}\ast W^k) + b^k
\end{equation}

The reconstruction is obtained by mapping back to the original image domain
with the latent maps $H$ and the decoding convolutional filter $\widetilde{W^k}$ : 
\begin{equation}
	\hat{\mathbf{x}} = \sigma(\sum_{k \in H} h^k \ast \widetilde{W^k} + c)
\end{equation}

where $\sigma:\mathbb{R} \to \mathbb{R}$ is a point-wise non-linearity like the sigmoid or
hyperbolic tangent function. A single bias value is broadcast to each component of a 
latent map. These $k$-output maps can be used as an input to the next layer of the CAE.
Several CAEs can be stacked into a deep hierarchy, which we again refer as a CAE to simplify 
the naming convention. 
We represent the stack of such operations as a single function 
$f_W : \mathbb{R}^{r \times c \times p} \to \mathbb{R}^{r \times c \times p}$
where the convolutional weights and biases are together represented by the weights $W$. 

In retrospect, the PCA, and traditional AE, ignore the spatial structure and location of pixels in the image.
This is also termed as being \textit{permutation invariant}. It is important to 
note that when working with image frames of few $100 \times 100$ pixels, these methods
introduce large redundancy in network parameters $W$, and furthermore span the entire visual 
receptive field. CAEs have fewer parameters on account of their weights being shared across many input locations/pixels.

\subsection{CAEs for Video anomaly detection}
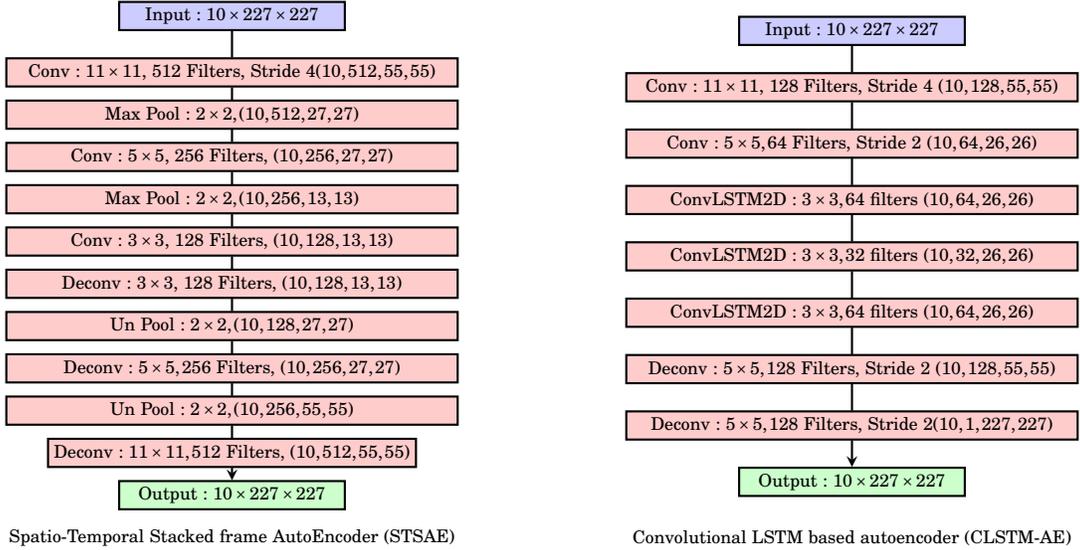
\begin{figure*}[htbp]
	\centering
	\begin{tikzpicture}[node distance=1cm, scale=0.75, every node/.style={transform shape}]
	\node[draw,thick,minimum width=4cm,minimum height=0.5cm, fill=blue!20] (in) at (1,0)  {Input : $10\times 227\times 227$};
	\node[draw,thick,minimum width=8cm,minimum height=0.5cm, fill=red!20] (conv1) at (1,-1)  {Conv : $11 \times 11$, $512$ Filters, Stride $4 (10,512,55,55)$};
	\node[draw,thick,minimum width=8cm,minimum height=0.5cm, fill=red!20] (pool1) at (1,-1.75)  {Max Pool : $2 \times 2, (10,512,27,27)$};
	\node[draw,thick,minimum width=8cm,minimum height=0.5cm, fill=red!20] (conv2) at (1,-2.5)  {Conv : $5 \times 5$, $256$ Filters, $ (10,256,27,27)$};
	\node[draw,thick,minimum width=8cm,minimum height=0.5cm, fill=red!20] (pool2) at (1,-3.25)  {Max Pool : $2 \times 2, (10,256,13,13)$}; 
	\node[draw,thick,minimum width=8cm,minimum height=0.5cm, fill=red!20] (conv3) at (1,-4)  {Conv : $3 \times 3$, $128$ Filters, $ (10,128,13,13)$};
	\node[draw,thick,minimum width=8cm,minimum height=0.5cm, fill=red!20] (deconv1) at (1,-4.75)  {Deconv : $3 \times 3$, $128$ Filters, $ (10,128,13,13)$};
	\node[draw,thick,minimum width=8cm,minimum height=0.5cm, fill=red!20] (unpool1) at (1,-5.5)  {Un Pool : $2 \times 2, (10,128,27,27)$}; 
	\node[draw,thick,minimum width=8cm,minimum height=0.5cm, fill=red!20] (deconv2) at (1,-6.25)  {Deconv : $5 \times 5, 256$ Filters, $(10,256,27,27)$};
	\node[draw,thick,minimum width=8cm,minimum height=0.5cm, fill=red!20] (unpool2) at (1,-7)  {Un Pool : $2 \times 2, (10,256,55,55)$}; 
	\node[draw,thick,minimum width=4cm,minimum height=0.5cm, fill=red!20] (deconv3) at (1,-7.75)  {Deconv : $11 \times 11, 512$ Filters, $(10,512,55,55)$};
	\node[draw,thick,minimum width=4cm,minimum height=0.5cm, fill=green!20] (out) at (1,-8.5)  {Output : $10\times 227\times 227$};
	\node at (1,-9.25) {Spatio-Temporal Stacked frame AutoEncoder (STSAE)};
	\draw [arrow] (in) -- (conv1) -- (pool1) -- (conv2) -- (pool2) -- (conv3) -- (deconv1) -- (unpool1) -- (deconv2) -- (unpool2) -- (deconv3) -- (out);
	\end{tikzpicture}
	\hspace{2cm}
	\begin{tikzpicture}[node distance=1cm, scale=0.75, every node/.style={transform shape}]
	\node[draw,thick,minimum width=4cm,minimum height=0.5cm, fill=blue!20] (in) at (1,0)  {Input : $10\times 227\times 227$};
	\node[draw,thick,minimum width=8cm,minimum height=0.5cm, fill=red!20] (conv1) at (1,-1)  {Conv : $11\times 11$, $128$ Filters, Stride $4$ $(10,128,55,55)$};
	\node[draw,thick,minimum width=8cm,minimum height=0.5cm, fill=red!20] (conv2) at (1,-2)  {Conv : $5 \times 5, 64$ Filters, Stride $2$ $(10,64,26,26)$};
	\node[draw,thick,minimum width=8cm,minimum height=0.5cm, fill=red!20] (lstm1) at (1,-3)  {ConvLSTM2D : $3 \times 3, 64$ filters $(10,64,26,26)$};
	\node[draw,thick,minimum width=8cm,minimum height=0.5cm, fill=red!20] (lstm2) at (1,-4)  {ConvLSTM2D : $3 \times 3, 32$ filters $(10,32,26,26)$};
	\node[draw,thick,minimum width=8cm,minimum height=0.5cm, fill=red!20] (lstm3) at (1,-5)  {ConvLSTM2D : $3 \times 3, 64$ filters $(10,64,26,26)$};
	\node[draw,thick,minimum width=8cm,minimum height=0.5cm, fill=red!20] (deconv1) at (1,-6)  {Deconv : $5 \times 5, 128$ Filters, Stride $2$ $(10,128,55,55)$};
	\node[draw,thick,minimum width=8cm,minimum height=0.5cm, fill=red!20] (deconv2) at (1,-7)  {Deconv : $5 \times 5, 128$ Filters, Stride $2 (10,1,227,227)$};
	\node[draw,thick,minimum width=4cm,minimum height=0.5cm, fill=green!20] (out) at (1,-8)  {Output : $10\times 227\times 227$};
	\node at (1,-9) {Convolutional LSTM based autoencoder (CLSTM-AE)};
	\draw [arrow] (in) -- (conv1) -- (conv2) -- (lstm1) -- (lstm2) -- (lstm3) -- (deconv1) -- (deconv2) -- (out);
	\end{tikzpicture}

	\caption{Spatio-Temporal Stacked frame AutoEncoder (left) : A sequence of 10 frames are being reconstructed by a convolutional autoencoder, 
	image reproduced from \cite{hasan2016learning}. Convolutional LSTM based autoencoder (right) : A sequence of 10 frames are being reconstructed by a spatio-temporal 
	autoencoder, 	\cite{chong2017abnormal}. The Convolutional LSTM layers are predictive models that
	model the spatio-temporal correlation of pixels in the video. This is described in 
	the predictive model section.}
	\label{fig:spatio-temporal-autoenc}
\end{figure*}

In the recent work by \cite{hasan2016learning}, a deep convolutional autoencoder was trained
to reconstruct an input sequence of frames from a training video set. We call this a 
Spatio-Temporal Stacked frame AutoEncoder (STSAE), to avoid confusion with similar names in the rest of the article. 
The STSAE in \cite{hasan2016learning} stacks $p$ frames $\mathbf{x}_i = [X_{i}, X_{i-1},..., X_{i-p+1}]$
with each time slice treated as a different channel in the input tensor to a convolutional autoencoder. 
The model is regularized by augmenting the loss function with L2-norm of the model weights :

\begin{equation}
	\mathcal{L}(W) = \frac{1}{2N} \sum_i \|\mathbf{x}_i - f_W(\mathbf{x}_i) \|_2^2 + \nu \|W\|_2^2
\end{equation}

where the tensor $\mathbf{x}_i \in \mathbb{R}^{r \times c \times p}$ is a cuboid with spatial dimensions $r, c$ are 
the spatial dimensions and $p$ is the number of frames temporally back into the past, with hyper-parameter $\nu$ which balances the reconstruction
error and norm of the parameters, and $N$ is the mini-batch size. The architecture of the convolutional autoencoder is reproduced 
in figure \ref{fig:spatio-temporal-autoenc}.

The image or tensor $\widehat{\mathbf{x}_i}$ reconstructed by the autoencoder enforces 
temporal regularity since the convolutional (weights) representation along with 
the bottleneck architecture of the autoencoder compresses information. The spatio-temporal autoencoder in 
\cite{chong2017abnormal} is shown in the right panel of figure \ref{fig:spatio-temporal-autoenc}. 
The reconstruction error map at frame $t$ is given by $E_t = |X_t-\hat{X}_t|$, while the 
temporal regularity score is given by the inverted, normalized reconstruction error : 

\begin{equation}
	s(t) = 1-\frac{\sum_{(x,y)}E_t-\min_{(x,y)}(E_t)}{\max_{(x,y)}(E_t)}
	\label{eqn:anomaly_score}
\end{equation}

where the $\sum$, $\min$ and $\max$ operators are across the spatial indices's $(x,y)$.
In other models, the normalized reconstruction error is directly used as the anomaly score.
One could envisage the use of Mahalanobis distance here since the task is to evaluate the distance between
test points from the points from the normal ones. This is evaluated as the error between the original tensor and 
the reconstruction from the autoencoder. 

Robust versions of Convolutional AutoEncoders (RCAE) are studied in \cite{chalapathy2017robustAE}, 
where the goal is to evaluate anomalies in images by imposing L2-constraints on parameters $W$
as well as adding a bias term. A video-patch (spatio-temporal) based autoencoder was employed 
by \cite{sabokrou2016video} to reconstruct patches, with a sparse autoencoder whose average 
activations were set to parameter $\rho$, enforcing sparseness, following the work in \cite{ng2011sparse}. 

\subsection{Contractive Autoencoders}

Contractive autoencoders explicitly create invariance by adding the Jacobian of the latent space
representation w.r.t the input of the autoencoder, to the reconstruction loss $L(\mathbf{x}, r(\mathbf{x}))$. This forces the latent space representation
to remain the same for small changes in the input \cite{rifai2011contractive}. Let us consider
the autoencoder with the encoder mapping the input image to the latent space $\mathbf{z} = f(\mathbf{x})$ 
and the decoder mapping back to the input image space $r(\mathbf{x}) = g(f(\mathbf{x}))$. The regularized loss function is : 

\begin{equation}
	\mathcal{L}(W) = \mathbb{E}_{\mathbf{x} \sim X_\text{train}} \bigg[L(\mathbf{x}, r(\mathbf{x})) + \lambda \norm{\frac{\partial f(\mathbf{x})}{\partial \mathbf{x}}}\bigg]
\end{equation}

Authors in \cite{alain2014regularized} describe what regularized autoencoders learn from the data generating the density function, 
and show for contractive and denoising encoders that this corresponds to the direction in which density is
increasing the most. Regularization forces the autoencoders to become less sensitive to input variation, 
though enforcing minimal reconstruction error keeps it sensitive to variations along
the manifold having high density. Contractive autoencoders capture variations on the manifold,
while mostly ignoring  variations orthogonal to it. Contractive autoencoder estimates the tangent
plane of the data manifold \cite{Goodfellow-et-al-2016}.

\subsection{Other deep models}

De-noising AutoEncoders (DAE) and Stacked DAEs (SDAEs) are well-known robust feature 
extraction methods in the domain of unsupervised learning \cite{vincent2008extracting},
where the reconstruction error minimization criteria is augmented with that of reconstructing from corrupted inputs.
SDAEs are used to learn representations from a video using both appearance, i.e. raw values, and motion information, 
i.e. optical flow between consecutive frames \cite{XuSebe2017}. Correlations between optical flow
and raw image values are modeled by coupling these two SDAE pipelines to learn a joint representation.

Deep belief networks (DBNs) are generative models, created by stacking multiple hidden layer units,
which are usually trained greedily to perform unsupervised feature learning. They are generative models
in the sense that they can reconstruct the original inputs. They have been discriminatively trained
using back-propagation \cite{erhan2010does} to achieve improved accuracies for supervised learning tasks.
The DBNs have been used to perform a raw image value based representation learning in \cite{Vu2017DBN}.

An early application of autoencoders to anomaly detection was performed in \cite{hawkins2002outlier}, 
on non-visual data. The Replicating neural network \cite{hawkins2002outlier}, constitutes of a  
feed-forward multi-layer perceptron with three hidden layers,
trained to map the training dataset to itself and anomalies correspond to large reconstruction error 
over test datasets. This is an autoencoder setup with a staircase like 
non-linearity applied at the middle hidden layer. The activation levels of this hidden units are thus quantized into
$N$ discrete values, $0, \frac{1}{N-1}, \frac{1}{N-2},..., 1$. The step-wise activation function used for the 
middle hidden layer divides the continuously distributed data points into a number of discrete-valued vectors.
The staircase non-linearity quantizes data points into clusters. This approach 
identifies cluster labels for each sample, and this often helps interpret resulting outliers. 

A rejection cascade over spatio-temporal cubes was generated to improve the performance speed of 
Deep-CNN based video anomaly detection framework by authors in  \cite{DeepCascade3D_DNN_2017}.

Videos can be viewed as a special case of spatio-temporal processes. A direct approach to video 
anomaly detection can be estimating the spatio-temporal mean and covariance. A major issue
is estimating the spatio-temporal covariance matrix due to its large size $n^2$ (where $n=$ $N$ pixels $\times$ $p$ frames).
In \cite{greenewald2014detection}, space-time pixel covariance for crowd videos were represented as a sum of Kronecker
products using only a few Kronecker factors, $\Sigma_{n \times n} \approxeq \sum_{i=1}^r \mathbf{T}_i \otimes \mathbf{S}_i$. 
To evaluate the anomaly score, the Mahanalobis distance for clips longer than 
the learned covariance needs to be evaluated. The inverse of the larger covariance matrix needs to be inferred from the estimated one, 
by block Toeplitz extension \cite{wiesel2013time}. It is to be noted that this study \cite{greenewald2014detection} only evaluates 
performance on the UMN dataset.

\section{Predictive modeling}

Predictive models aim to model the current output frame $X_t$ as a function of the past $p$ frames $[ X_{t-1}, X_{t-2}, ..., X_{t-p+1}]$.
This is well-known in time series analysis under auto-regressive models, where the function over the past is linear. 
Recurrent neural networks (RNN) model this function as a recurrence relationship, frequently involving a non-linearity such as a sigmoid function.
LSTM is the standard model for sequence prediction. It learns a gating function over the classical RNN architecture to prevent the vanishing gradient problem during backpropagation through time (BPTT) \cite{hochreiter1997long}. 
Recently there have also been attempts to perform efficient video prediction using feed-forward convolutional networks for 
video prediction by minimizing the mean-squared error (MSE) between predicted and future frames \cite{mathieu2015deep}.  
Similar efforts were performed in \cite{lotter2015unsupervised} using a CNN-LSTM-deCNN framework while combining MSE and an
adversarial loss.

\subsection{Composite Model : Reconstruction and prediction}

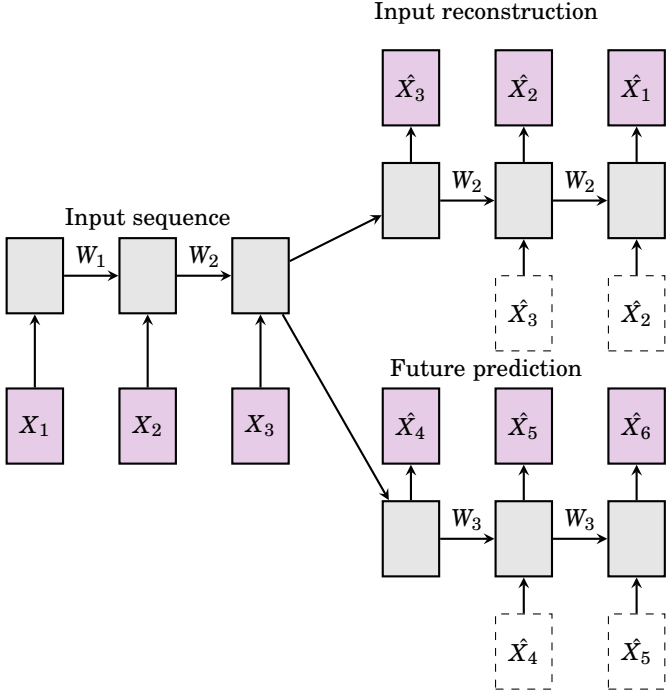
\begin{figure}[htbp]
	\centering
	\begin{tikzpicture}[node distance=1cm]
	\node[draw,thick,minimum width=0.75cm,minimum height=1cm,fill=violet!20] (x1) at (1,0)  {$X_1$};
	\node[draw,thick,minimum width=0.75cm,minimum height=1cm,fill=violet!20] (x2) at (2.5,0)  {$X_2$};
	\node[draw,thick,minimum width=0.75cm,minimum height=1cm,fill=violet!20] (x3) at (4,0)  {$X_3$};

	\node[draw,thick,minimum width=0.75cm,minimum height=1cm,fill=gray!20] (h1) at (1,2)  {};
	\node[draw,thick,minimum width=0.75cm,minimum height=1cm,fill=gray!20] (h2) at (2.5,2)  {};
	\node[draw,thick,minimum width=0.75cm,minimum height=1cm,fill=gray!20] (h3) at (4,2)  {};

	\node[draw,thick,minimum width=0.75cm,minimum height=1cm,fill=gray!20] (h1rec) at (6,3)   {};
	\node[draw,thick,minimum width=0.75cm,minimum height=1cm,fill=gray!20] (h2rec) at (7.5,3) {};
	\node[draw,thick,minimum width=0.75cm,minimum height=1cm,fill=gray!20] (h3rec) at (9,3)  {};
	\node[draw,dashed,minimum width=0.75cm,minimum height=1cm,fill=white] (x1recroll) at (7.5,1.5) {$\hat{X_3}$};
	\node[draw,dashed,minimum width=0.75cm,minimum height=1cm,fill=white] (x2recroll) at (9,1.5) {$\hat{X_2}$};

	\node[draw,thick,minimum width=0.75cm,minimum height=1cm,fill=violet!20] (x1rec) at (6,4.5)   {$\hat{X_3}$};
	\node[draw,thick,minimum width=0.75cm,minimum height=1cm,fill=violet!20] (x2rec) at (7.5,4.5) {$\hat{X_2}$};
	\node[draw,thick,minimum width=0.75cm,minimum height=1cm,fill=violet!20] (x3rec) at (9,4.5)  {$\hat{X_1}$};

	\node[draw,thick,minimum width=0.75cm,minimum height=1cm,fill=gray!20] (h1pred) at (6,-1.5)   {};
	\node[draw,thick,minimum width=0.75cm,minimum height=1cm,fill=gray!20] (h2pred) at (7.5,-1.5) {};
	\node[draw,thick,minimum width=0.75cm,minimum height=1cm,fill=gray!20] (h3pred) at (9,-1.5)  {};
	\node[draw,dashed,minimum width=0.75cm,minimum height=1cm,fill=white] (x1predroll) at (7.5,-3) {$\hat{X_4}$};
	\node[draw,dashed,minimum width=0.75cm,minimum height=1cm,fill=white] (x2predroll) at (9,-3) {$\hat{X_5}$};

	\node[draw,thick,minimum width=0.75cm,minimum height=1cm,fill=violet!20] (x4pred) at (6,0)   {$\hat{X_4}$};
	\node[draw,thick,minimum width=0.75cm,minimum height=1cm,fill=violet!20] (x5pred) at (7.5,0) {$\hat{X_5}$};
	\node[draw,thick,minimum width=0.75cm,minimum height=1cm,fill=violet!20] (x6pred) at (9,0)  {$\hat{X_6 }$};

	\node at (7,5.5) {Input reconstruction};
	\node at (7,0.75) {Future prediction};
	\node at (2.5,2.75) {Input sequence};

	\draw[arrow] (x1) -- (h1);
	\draw[arrow] (x2) -- (h2);
	\draw[arrow] (x3) -- (h3);
	\draw[arrow] (h1) -- (h2) node[midway,above]{$W_1$};
	\draw[arrow] (h2) -- (h3) node[midway,above]{$W_2$};
	\draw[arrow] (h3) -- (h1rec);
	\draw[arrow] (h3) -- (h1pred);
	\draw[arrow] (h1rec) -- (x1rec);
	\draw[arrow] (h2rec) -- (x2rec);
	\draw[arrow] (h3rec) -- (x3rec);
	\draw[arrow] (h1pred) -- (x4pred);
	\draw[arrow] (h2pred) -- (x5pred);
	\draw[arrow] (h3pred) -- (x6pred);
	\draw[arrow] (x1recroll) -- (h2rec);
	\draw[arrow] (x2recroll) -- (h3rec);
	\draw[arrow] (x1predroll) -- (h2pred);
	\draw[arrow] (x2predroll) -- (h3pred);
	\draw[arrow] (h1rec) -- (h2rec) node[midway,above]{$W_2$};
	\draw[arrow] (h2rec) -- (h3rec) node[midway,above]{$W_2$};
	\draw[arrow] (h1pred) -- (h2pred) node[midway,above]{$W_3$};
	\draw[arrow] (h2pred) -- (h3pred) node[midway,above]{$W_3$};
	\end{tikzpicture}

	\caption{Composite LSTM module \cite{srivastava2015unsupervisedLSTM}.
	The LSTM model weights $W$ represent fully connected layers.}
	\label{fig:compositelstm}
\end{figure}

This composite LSTM model in \cite{srivastava2015unsupervisedLSTM}, 
combines an autoencoder model and predictive LSTM model, see figure \ref{fig:compositelstm}.
Autoencoders suffer learning trivial representations of input, 
by memorization, while memorization is not useful for predicting future frames.  
On the other hand, the future predictor's role requires memory of
temporally past few frames, though this would not be compatible with the autoencoder
loss which is more global. The composite model was used to extract features
from video data for the tasks of action recognition. The composite LSTM model is defined using a 
fully connected LSTM (FC-LSTM) layer.

\subsection{Convolutional LSTM}

Convolutional long short-term memory (ConvLSTM) model \cite{xingjian2015convolutionalLSTM} 
is a composite LSTM based encoder-decoder model. 
FC-LSTM does not take spatial correlation into consideration and is permutation invariant to pixels, 
while a ConvLSTM has convolutional layers instead of fully connected layers, thus modeling 
spatio-temporal correlations. The ConvLSTM as described in equations \ref{eqn:convlstm} evaluates 
future states of cells in a spatial grid as a function of the inputs and past states of its local neighbors. 

Authors in \cite{xingjian2015convolutionalLSTM}, consider a spatial grid, with each grid 
cell containing multiple spatial measurements, which they aim to forecast for the next $K$ 
future frames, given $J$ observations in the past. The spatio-temporal
correlations are used as input to a recurrent model, the convolutional LSTM. The equations for input, gating and the output are presented below. 

\begin{equation}
	\begin{split}
	    i_t = \ & \sigma(W_{xi} \ast X_t + W_{hi} \ast H_{t-1} + W_{ci} \circ C_{t-1} + b_i) \\
	    f_t = \ & \sigma(W_{xf} \ast X_t + W_{hf} \ast H_{t-1} + W_{cf} \circ C_{t-1} + b_f) \\
	    C_t = \ & f_t \circ C_{t-1} + i_t \circ \text{tanh}(W_{xc}+X_t+W_{hc} \ast H_{t-1}+b_c) \\
	    o_t = \ & \sigma(W_{xo} \ast X_t + W_{ho} \ast H_{t-1} + W_{co} \circ C_{t} + b_o) \\
	    H_t = \ & o_t \circ \text{tanh}(C_t) \\
	\end{split}
	\label{eqn:convlstm}
\end{equation}
Here, $\ast$ refers to the convolution operation while $\circ$ refers to the Hadamard product, the element-wise product of matrices.
Encoding network compresses the input sequence into a hidden state tensor while the forecasting network 
unfolds the hidden state tensor to make a prediction. The hidden representations can be used to represent moving 
objects in the scene, a larger transitional kernel captures  
faster motions compared to smaller kernels \cite{xingjian2015convolutionalLSTM}.

\begin{figure}[htbp]
	\centering
	\begin{tikzpicture}[scale=0.8, every node/.style={transform shape}]
	\tikzset{
	arrow/.style={
	            color=black,
	            draw=blue,
	            -latex,
	                font=\fontsize{8}{8}\selectfont},
	        }

	\pgfmathsetmacro{\cubex}{4}
	\pgfmathsetmacro{\cubey}{0.25}
	\pgfmathsetmacro{\cubez}{4}
	\draw[black,fill=yellow, fill opacity=0.2] (0,0,0) -- ++(-\cubex,0,0) -- ++(0,-\cubey,0) -- ++(\cubex,0,0) -- cycle;
	\draw[black,fill=yellow, fill opacity=0.2] (0,0,0) -- ++(0,0,-\cubez) -- ++(0,-\cubey,0) -- ++(0,0,\cubez) -- cycle;
	\draw[black,fill=yellow, fill opacity=0.2] (0,0,0) -- ++(-\cubex,0,0) -- ++(0,0,-\cubez) -- ++(\cubex,0,0) -- cycle;

	\pgfmathsetmacro{\cubex}{4}
	\pgfmathsetmacro{\cubey}{0.5}
	\pgfmathsetmacro{\cubez}{4}
	\draw[black,fill=yellow, fill opacity=0.2] (0,2,0) -- ++(-\cubex,0,0) -- ++(0,-\cubey,0) -- ++(\cubex,0,0) -- cycle;
	\draw[black,fill=yellow, fill opacity=0.2] (0,2,0) -- ++(0,0,-\cubez) -- ++(0,-\cubey,0) -- ++(0,0,\cubez) -- cycle;
	\draw[black,fill=yellow, fill opacity=0.2] (0,2,0) -- ++(-\cubex,0,0) -- ++(0,0,-\cubez) -- ++(\cubex,0,0) -- cycle;

	\draw[arrow, thick](-4,1.5,0) to [bend left,looseness=1.5] node[] {} (-3,1.5,0);
	\draw[arrow, thick](-4,3.5,0) to [bend left,looseness=1.5] node[] {} (-3,3.5,0);
	\draw[arrow, thick](3,1.5,0) to [bend left,looseness=1.5] node[] {} (4,1.5,0);
	\draw[arrow, thick](3,3.5,0) to [bend left,looseness=1.5] node[] {} (4,3.5,0);

	\draw[arrow, thick, red](6,3.5,0) to [bend right,looseness=1.5] node[] {} (6,6,0);
	\draw[arrow, thick, red](6,1.5,0) to [bend right,looseness=1.5] node[] {} (6,6,0);

	\pgfmathsetmacro{\cubex}{2}
	\pgfmathsetmacro{\cubey}{0.05}
	\pgfmathsetmacro{\cubez}{2}
	\draw[black,fill=blue, fill opacity=0.4] (-1,0,0) -- ++(-\cubex,0,0) -- ++(0,-\cubey,0) -- ++(\cubex,0,0) -- cycle;
	\draw[black,fill=blue, fill opacity=0.4] (-1,0,0) -- ++(0,0,-\cubez) -- ++(0,-\cubey,0) -- ++(0,0,\cubez) -- cycle;
	\draw[black,fill=blue, fill opacity=0.4] (-1,0,0) -- ++(-\cubex,0,0) -- ++(0,0,-\cubez) -- ++(\cubex,0,0) -- cycle;

	\draw (-1,0,0) -- (-2,2.25,0);
	\draw (-3,0,0) -- (-2,2.25,0);
	\draw (-3,0,-2) -- (-2,2.25,0);
	\draw (-1,0,-2) -- (-2,2.25,0);

	\pgfmathsetmacro{\cubex}{4}
	\pgfmathsetmacro{\cubey}{0.5}
	\pgfmathsetmacro{\cubez}{4}
	\draw[black,fill=yellow, fill opacity=0.2] (0,4,0) -- ++(-\cubex,0,0) -- ++(0,-\cubey,0) -- ++(\cubex,0,0) -- cycle;
	\draw[black,fill=yellow, fill opacity=0.2] (0,4,0) -- ++(0,0,-\cubez) -- ++(0,-\cubey,0) -- ++(0,0,\cubez) -- cycle;
	\draw[black,fill=yellow, fill opacity=0.2] (0,4,0) -- ++(-\cubex,0,0) -- ++(0,0,-\cubez) -- ++(\cubex,0,0) -- cycle;

	\pgfmathsetmacro{\cubex}{2}
	\pgfmathsetmacro{\cubey}{0.05}
	\pgfmathsetmacro{\cubez}{2}
	\draw[black,fill=blue, fill opacity=0.4] (-1,2,0) -- ++(-\cubex,0,0) -- ++(0,-\cubey,0) -- ++(\cubex,0,0) -- cycle;
	\draw[black,fill=blue, fill opacity=0.4] (-1,2,0) -- ++(0,0,-\cubez) -- ++(0,-\cubey,0) -- ++(0,0,\cubez) -- cycle;
	\draw[black,fill=blue, fill opacity=0.4] (-1,2,0) -- ++(-\cubex,0,0) -- ++(0,0,-\cubez) -- ++(\cubex,0,0) -- cycle;

	\draw (-1,2,0) -- (-2,4.25,0);
	\draw (-3,2,0) -- (-2,4.25,0);
	\draw (-3,2,-2) -- (-2,4.25,0);
	\draw (-1,2,-2) -- (-2,4.25,0);

	\draw[red,thick,->] (0,4.25,0) -- (3,4.25,0) node[midway,above]{copy};
	\draw[red,thick,->] (0,2.25,0) -- (3,2.25,0) node[midway,above]{copy};

	\node[] at (0,1,0) {Input};
	\node[] at (0,3.2,0) {ConvLSTM$_1$};
	\node[] at (0,5.2,0) {ConvLSTM$_2$};
	\node[] at (5,3.2,0) {ConvLSTM$_3$};
	\node[] at (5,5.2,0) {ConvLSTM$_4$};
	\node[] at (0,6.2,0) {\bf Encoding n/w};
	\node[] at (5,7.2,0) {Prediction};
	\node[] at (5,1,0) {\bf Forecasting n/w};

	\pgfmathsetmacro{\cubex}{2}
	\pgfmathsetmacro{\cubey}{0.05}
	\pgfmathsetmacro{\cubez}{2}
	\draw[black,fill=blue, fill opacity=0.4] (5,2,0) -- ++(-\cubex,0,0) -- ++(0,-\cubey,0) -- ++(\cubex,0,0) -- cycle;
	\draw[black,fill=blue, fill opacity=0.4] (5,2,0) -- ++(0,0,-\cubez) -- ++(0,-\cubey,0) -- ++(0,0,\cubez) -- cycle;
	\draw[black,fill=blue, fill opacity=0.4] (5,2,0) -- ++(-\cubex,0,0) -- ++(0,0,-\cubez) -- ++(\cubex,0,0) -- cycle;

	\draw (5,2,0) -- (4,4.25,0);
	\draw (3,2,0) -- (4,4.25,0);
	\draw (3,2,-2) -- (4,4.25,0);
	\draw (5,2,-2) -- (4,4.25,0);

	\pgfmathsetmacro{\cubex}{4}
	\pgfmathsetmacro{\cubey}{0.5}
	\pgfmathsetmacro{\cubez}{4}
	\draw[black,fill=yellow, fill opacity=0.2] (6,2,0) -- ++(-\cubex,0,0) -- ++(0,-\cubey,0) -- ++(\cubex,0,0) -- cycle;
	\draw[black,fill=yellow, fill opacity=0.2] (6,2,0) -- ++(0,0,-\cubez) -- ++(0,-\cubey,0) -- ++(0,0,\cubez) -- cycle;
	\draw[black,fill=yellow, fill opacity=0.2] (6,2,0) -- ++(-\cubex,0,0) -- ++(0,0,-\cubez) -- ++(\cubex,0,0) -- cycle;

	\pgfmathsetmacro{\cubex}{4}
	\pgfmathsetmacro{\cubey}{0.5}
	\pgfmathsetmacro{\cubez}{4}
	\draw[black,fill=yellow, fill opacity=0.2] (6,4,0) -- ++(-\cubex,0,0) -- ++(0,-\cubey,0) -- ++(\cubex,0,0) -- cycle;
	\draw[black,fill=yellow, fill opacity=0.2] (6,4,0) -- ++(0,0,-\cubez) -- ++(0,-\cubey,0) -- ++(0,0,\cubez) -- cycle;
	\draw[black,fill=yellow, fill opacity=0.2] (6,4,0) -- ++(-\cubex,0,0) -- ++(0,0,-\cubez) -- ++(\cubex,0,0) -- cycle;

	\pgfmathsetmacro{\cubex}{4}
	\pgfmathsetmacro{\cubey}{0.25}
	\pgfmathsetmacro{\cubez}{4}
	\draw[red,fill=yellow, fill opacity=0.2] (6,6,0) -- ++(-\cubex,0,0) -- ++(0,-\cubey,0) -- ++(\cubex,0,0) -- cycle;
	\draw[red,fill=yellow, fill opacity=0.2] (6,6,0) -- ++(0,0,-\cubez) -- ++(0,-\cubey,0) -- ++(0,0,\cubez) -- cycle;
	\draw[red,fill=yellow, fill opacity=0.2] (6,6,0) -- ++(-\cubex,0,0) -- ++(0,0,-\cubez) -- ++(\cubex,0,0) -- cycle;

	\end{tikzpicture}
	\caption{A convolutional LSTM architecture for spatio-temporal prediction.}
	\label{fig:convlstm_arch}
\end{figure}
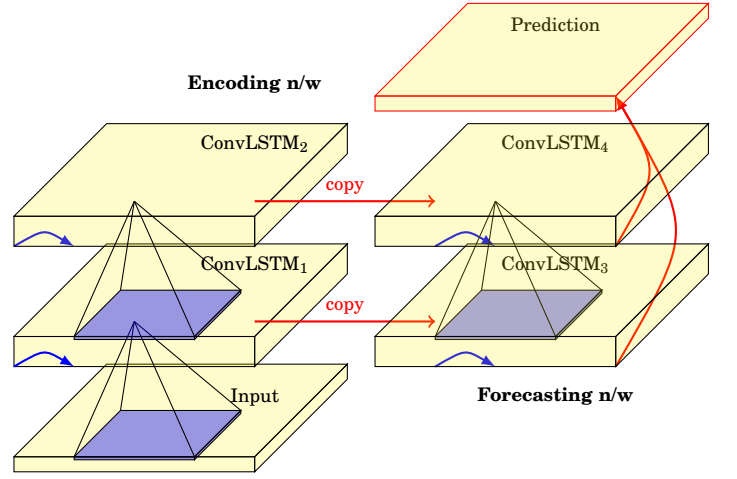

The ConvLSTM model was used as a unit within the composite LSTM model 
\cite{srivastava2015unsupervisedLSTM} following an encoder-decoder, 
with a branch for reconstruction and another for prediction. This 
architecture was applied for video anomaly detection by 
\cite{medel2016anomalythesis}, \cite{medel2016anomaly}, with promising results.

In \cite{luo2017remembering}, a convolutional representation of the input video is used 
as input to the convolutional LSTM and a de-convolution to reconstruct the ConvLSTM output 
to the original resolution. The authors call this a ConvLSTM Autoencoder, though fundamentally
it is not very different from a ConvLSTM.

\subsection{3D-Autoencoder and Predictor}

As remarked by authors in \cite{Zhao20173DAE_STAE}, while 2D-ConvNets are appropriate 
representations learnt for image recognition and detection tasks, they are 
incapable of capturing the temporal information encoded in consecutive frames 
for video analysis problems. 3-D convolutional architectures are known to perform 
well for action recognition \cite{tran2015learning_C3D}, and are used in the form 
of an autoencoder. Such a 3D autoencoder learns representations that are
invariant to spatio-temporal changes (movement) encoded by the 3-D convolutional 
feature maps. Authors in \cite{Zhao20173DAE_STAE} propose to use a 3D kernel by 
stacking T-frames together as in \cite{hasan2016learning}. The output feature map 
of each kernel is a 3D tensor including the temporal dimension and are aimed 
to summarize motion information. 

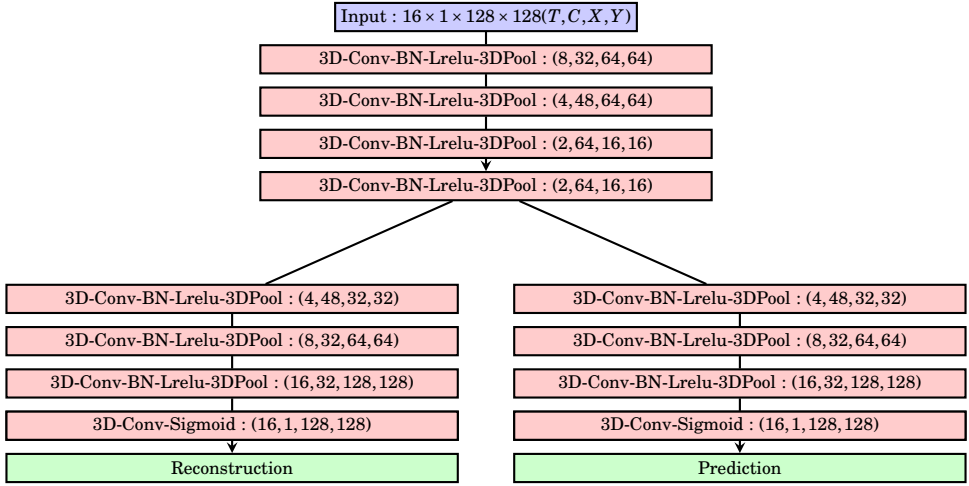
\begin{figure*}[htbp]
	\centering
	\begin{tikzpicture}[node distance=1cm, scale=0.75, every node/.style={transform shape}]
	\node[draw,thick,minimum width=4cm,minimum height=0.5cm, fill=blue!20] (in) at (2.5,0)  {Input : $16 \times 1 \times 128 \times 128 (T,C,X,Y)$};
	\node[draw,thick,minimum width=8cm,minimum height=0.5cm, fill=red!20] (3dconv1) at (2.5,-0.75)  {3D-Conv-BN-Lrelu-3DPool : $(8,32,64,64)$};
	\node[draw,thick,minimum width=8cm,minimum height=0.5cm, fill=red!20] (3dconv2) at (2.5,-1.5)  {3D-Conv-BN-Lrelu-3DPool : $(4,48,64,64)$};
	\node[draw,thick,minimum width=8cm,minimum height=0.5cm, fill=red!20] (3dconv3) at (2.5,-2.25)  {3D-Conv-BN-Lrelu-3DPool : $(2,64,16,16)$};
	\node[draw,thick,minimum width=8cm,minimum height=0.5cm, fill=red!20] (3dconv4) at (2.5,-3)  {3D-Conv-BN-Lrelu-3DPool : $(2,64,16,16)$};

	\node[draw,thick,minimum width=8cm,minimum height=0.5cm, fill=red!20] (3dconv1recon) at (-2,-5)  {3D-Conv-BN-Lrelu-3DPool : $(4,48,32,32)$};
	\node[draw,thick,minimum width=8cm,minimum height=0.5cm, fill=red!20] (3dconv2recon) at (-2,-5.75)  {3D-Conv-BN-Lrelu-3DPool : $(8,32,64,64)$};
	\node[draw,thick,minimum width=8cm,minimum height=0.5cm, fill=red!20] (3dconv3recon) at (-2,-6.5)  {3D-Conv-BN-Lrelu-3DPool : $(16,32,128,128)$};
	\node[draw,thick,minimum width=8cm,minimum height=0.5cm, fill=red!20] (3dconv4recon) at (-2,-7.25)  {3D-Conv-BN-Lrelu-3DPool : $(16,32,128,128)$};
	\node[draw,thick,minimum width=8cm,minimum height=0.5cm, fill=red!20] (recon) at (-2,-7.25)  {3D-Conv-Sigmoid : $(16,1,128,128)$};
	\node[draw,thick,minimum width=8cm,minimum height=0.5cm, fill=green!20] (recon) at (-2,-8)  {Reconstruction};

	\node[draw,thick,minimum width=8cm,minimum height=0.5cm, fill=red!20] (3dconv1pred) at (7,-5)  {3D-Conv-BN-Lrelu-3DPool : $(4,48,32,32)$};
	\node[draw,thick,minimum width=8cm,minimum height=0.5cm, fill=red!20] (3dconv2pred) at (7,-5.75)  {3D-Conv-BN-Lrelu-3DPool : $(8,32,64,64)$};
	\node[draw,thick,minimum width=8cm,minimum height=0.5cm, fill=red!20] (3dconv3pred) at (7,-6.5)  {3D-Conv-BN-Lrelu-3DPool : $(16,32,128,128)$};
	\node[draw,thick,minimum width=8cm,minimum height=0.5cm, fill=red!20] (3dconv4pred) at (7,-7.25)  {3D-Conv-BN-Lrelu-3DPool : $(16,32,128,128)$};
	\node[draw,thick,minimum width=8cm,minimum height=0.5cm, fill=red!20] (pred) at (7,-7.25)  {3D-Conv-Sigmoid : $(16,1,128,128)$};
	\node[draw,thick,minimum width=8cm,minimum height=0.5cm, fill=green!20] (pred) at (7,-8)  {Prediction};

	\draw [arrow] (in) -- (3dconv1) -- (3dconv2) -- (3dconv3) -- (3dconv4); 
	\draw [arrow] (3dconv4) -- (3dconv1recon) -- (3dconv2recon) -- (3dconv3recon) -- (3dconv4recon) -- (recon);
	\draw [arrow] (3dconv4) -- (3dconv1pred) -- (3dconv2pred) -- (3dconv3pred) -- (3dconv4pred) -- (pred);
	\end{tikzpicture}

	\caption{Spatio-temporal autoencoder architecture from \cite{Zhao20173DAE_STAE} with 
	reconstruction and prediction branches, following the composite model in \cite{srivastava2015unsupervisedLSTM}. 
	Batch Normalization(BN) is applied at each layer, following which
	a leaky Relu non-linearity is applied, finally followed by a 3D max-pooling operation.}
	\label{fig:STAE_recon_prediction}
\end{figure*}

The reconstruction branch follows an autoencoder loss: 

\begin{equation}
	\mathcal{L}_{\text{rec}}(W)=\frac{1}{N} \sum_{i=1}^N \|X_i-f_\text{rec}^W(X_i)\|_2^2
\end{equation}

The prediction branch loss is inversely weighted by moving window's length that falls 
off symmetrically w.r.t the current frame, to reduce
the effect of past frames on the predicted frame : 

\begin{equation}
	\mathcal{L}_\text{pred}(W) = \frac{1}{N} \sum_{i=1}^N \frac{1}{T^2} \sum_{t=1}^T (T-t) \|X_{i+T}-f_\text{pred}^W(X_i)\|_2^2
\end{equation}

Thus the final optimization objective minimized is :

\begin{equation}
	\mathcal{L}(W) = \mathcal{L}_\text{rec}(W) + \mathcal{L}_\text{pred}(W) + \lambda \|W\|^2_2
\end{equation}

Anomalous regions where spatio-temporal blocks, that even when poorly reconstructed by the autoencoder branch, would be well
predicted by the prediction branch. The prediction loss was designed to enforce local temporal coherence by tracking spatio-temporal correlation,
and not for the prediction of the appearance of new objects in the relatively long-term future. 

\subsection{Slow Feature Analysis (SFA)}

Slow feature analysis \cite{wiskott2002slow} is an unsupervised representation 
learning method which aims at extracting slowly varying representations of 
rapidly varying high dimensional input. The SFA is based on the \textit{slowness principle},
which states that the responses of individual receptors or pixel variations 
are highly sensitive to local variations in the environment, and thus vary 
much faster, while the higher order internal visual representations vary on a slow
timescale. From a predictive modeling perspective SFA extracts a representation 
$\mathbf{y}(t)$ of the high dimensional input $\mathbf{x}_t$ that maximizes 
information on the next time sample $\mathbf{x}_{t+1}$. Given a high dimensional input
varying over time, $[\mathbf{x}_1, \mathbf{x}_2, ..., \mathbf{x}_T], t \in [t_0, t_1]$ 
SFA extracts a representation $\mathbf{y} = f_\theta(\mathbf{x})$ which is a solution 
to the following optimization problem \cite{creutzig2008predictive} :

\begin{equation}
	 \argmin_{\mathbf{E}_t [\mathbf{y}_{t}] = 0, \ \ \mathbf{E}_t [y_j^2]=1, \ \ \mathbf{E}_t[y_{j}y_{j^\prime}]=0] = 1,} \mathbf{E}_t [\mathbf{y}_{t+1}-\mathbf{y}_{t}]	
\end{equation}

As seen in the constraints, the representation is enforced to have, zero mean to  
ensure a unique solution, while unit covariance to avoid trivial zero solution. Feature de-correlation
removes redundancy across the features.

SFA has been well known in pattern recognition and has been applied to the problem of  
activity recognition \cite{sun2014dlsfa}, \cite{zhang2012slow}. Authors in \cite{kompella2012incrementalSFA} 
propose an incremental application of SFA that updates slow features incrementally. The SFA is 
calculated using batch PCA, iterated twice. The first PCA to whiten the inputs. 
The second PCA is applied on the derivative of the normalized input to evaluate 
the flow features. To achieve a computationally tractable solution, a two-layer 
localized SFA architecture is proposed by authors \cite{hu2016videoSFA} for the 
task of online slow feature extraction and consequent anomaly detection.

\noindent \textbf{Other Predictive models} :
A convolutional feature representation was fed into an LSTM model to predict the latent 
space representation and its prediction error was used to evaluate anomalies in a 
robotics application \cite{munawar2017spatiotemporalAnomaly}. A recurrent 
autoencoder using an LSTM that models temporal dependence between patches from a 
sequence of input frames is used to detect video forgery \cite{d2017autoencoder}. 

\section{Deep generative Models}

\subsection{Generative Vs Discriminative} 

Let us consider a supervised learning setup $(X_i, y_i) \in \mathbb{R}^{d} \times \{C_j\}_{j=1}^K$, where $i$
indexes the number of samples $i=1:N$ in the dataset. 
Generative models estimate class conditional posterior distribution $P(X|y)$, which can be difficult
if the input data are high dimensional images or spatio-temporal tensors.  A discriminative 
model evaluates the class probability $P(y|X)$ directly from the data to classify the samples $X$
into different classes $\{C_j\}_{j=1}^K$. 

Deep generative models that can learn via the principle of maximum likelihood differ 
with respect to how they represent or approximate the likelihood.
The explicit models are ones where the density $p_\text{model}(\mathbf{x}, \boldsymbol{\theta})$
is evaluated explicitly and the likelihood maximized.

In this section, we will review the stochastic autoencoders; the variational autoencoder
and the adversarial autoencoder, and their applications to the problem of anomaly detection. 
And finally the generative adversarial networks to anomaly detection in images and videos.

\subsection{Variational Autoencoders (VAEs)}
Variational Autoencoders \cite{kingmaVAE2014ICLR} are generative models that approximate the data distribution $P(X)$ 
of a high dimensional input $X$, an image or video. Variational approximation of the latent space
is achieved using an autoencoder architecture, with a probabilistic encoder $q_\phi(\mathbf{x}|\mathbf{z})$ that produces Gaussian distribution
in the latent space, and a probabilistic decoder $p_\theta(\mathbf{z}|\mathbf{x})$, which given a code produces distribution
over the input space. The motivation behind variational methods is to 
pick a family of distributions over the latent variables with its own variational parameters $q_{\phi}(\mathbf{z})$, 
and estimate the parameters for this family so that it approaches $q_{\phi}$.

The loss function constitutes of the KL-Divergence regularization term, and the expected negative reconstruction error with 
an additional KL-divergence term between the latent space vector and the representation with 
a mean vector and a standard deviation vector, that optimizes the variational lower bound on the 
marginal log-likelihood of each observation. 

\begin{equation}
\label{eqn:vae_loss}
	\begin{split}
		& \mathcal{L}(\theta, \phi, \mathbf{x}^{(i)}) = \sum_{i=1}^{N_{\text{train}}} -D_{KL}(q_{\phi}(\mathbf{z}|\mathbf{x}^{(i)})||p_{\theta}(\mathbf{z}))
		+ \frac{1}{L} \sum_{l=1}^L (\log p_{\theta}(\mathbf{x}^{(i)}|\mathbf{z}^{(i,l)}))\\
		& \text{where \ } \mathbf{z}^{(i,l)} = g_\phi(\boldsymbol{\epsilon}^{(i,l)}, \mathbf{x}^{(i)})
		\text{ \ and \ } \boldsymbol{\epsilon}^{(l)} \sim p(\boldsymbol{\epsilon})
	\end{split}
\end{equation}

The function $g_\phi$ maps sample $\mathbf{x}^{(i)}$ and noise vector $\boldsymbol{\epsilon}^{(l)}$
to a sample from the approximate posterior for that data-point 
$\mathbf{z}^{(i,l)} = g_\phi(\boldsymbol{\epsilon}^{(l)}, \mathbf{x}^{(i)})$ where,
$\mathbf{z}^{(i,l)} \sim q_\phi(\mathbf{z}|\mathbf{x}^{(i)})$. To solve this sampling problem
authors \cite{kingmaVAE2014ICLR} propose the reparameterization trick. The random variable 
$\mathbf{z} \sim q_\phi(\mathbf{z}|\mathbf{x})$ is expressed as function of a deterministic variable
$\mathbf{z}=g_\phi(\boldsymbol{\epsilon}, \mathbf{z})$ where $\boldsymbol{\epsilon}$ is an auxiliary variable
with independent marginal $p(\boldsymbol{\epsilon})$. This reparameterization rewrites an expectation w.r.t
$q_\phi(\mathbf{z}|\mathbf{x})$ such that the Monte Carlo estimate of the expectation is differentiable w.r.t.
$\phi$. A valid reparameterization was the unit-Gaussian 
case $\mathbf{z}^{(i,l)} = \boldsymbol{\mu}^{(i,l)} + \sigma^{(i)} \odot \boldsymbol{\epsilon}^{(l)} $
where $\boldsymbol{\epsilon}^{(l)} \sim \mathcal{N}(0,I)$.

Specifically for a VAE, the goal is to learn a low dimensional representation $z$
by modeling $p_\theta(\mathbf{x}|\mathbf{z})$ with a simpler distribution, a 
centered isotropic multivariate Gaussian, i.e. $p_\theta(\mathbf{z}) = \mathcal{N}(\mathbf{z},\mathbf{0},I)$. 
In this model both the prior $p_\theta(\mathbf{z})$, and $q_\phi(\mathbf{z}|\mathbf{x})$ are Gaussian; and the resulting loss function was described in equation \ref{eqn:vae_loss}.

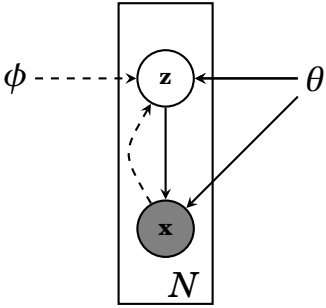
\begin{figure}[htbp]
	\centering
	
	\begin{tikzpicture}
	\node[draw,thick,minimum width=1.25cm,minimum height=4cm] at (1,0)  {};
	\node[draw, thick, circle, minimum width=0.75cm] (z) at (1,1) {$\mathbf{z}$};
	\node[draw, thick, circle, minimum width=0.75cm, fill=gray] (x) at (1,-1) {$\mathbf{x}$};
	\node (phi) at (-1,1) {\Large $\phi$};
	\node (theta) at (3,1) {\Large $\theta$};
	\node (N) at (1.25,-1.75) {\Large $N$};

	\draw[arrow] (z)--(x);
	\draw[arrow,dashed] (phi)--(z);
	\draw[arrow,dashed] (x) to [bend left,looseness=1.5] (z);
	\draw[arrow] (theta)--(z);
	\draw[arrow] (theta)--(x);
	\draw[arrow] (theta)--(z);
	\end{tikzpicture}
	\caption{Graphical model for the VAE : Solid lines denote the generative model $p_{\theta}(\mathbf{z})p_{\theta}(\mathbf{x}|\mathbf{z})$, dashed lines denote the variational approximation $q_\phi(\mathbf{z}|\mathbf{x})$ to the intractable posterior $p_\theta(\mathbf{z}|\mathbf{x})$  \cite{kingmaVAE2014ICLR}.}
	\label{fig:graphicalVAE}
\end{figure}

\subsection{Anomaly detection using VAE}
Anomaly detection using the VAE framework has been studied in \cite{an2015variational}. 
Authors define the reconstruction probability as $\mathbb{E}_{q_{\phi}(\mathbf{z}|\mathbf{x})} [\log p_{\theta}(\mathbf{x}|\mathbf{z})]$.
Once the VAE is trained, for a new test sample $\mathbf{x}^{(i)}$, one first evaluates 
the mean and standard deviation vectors with the probabislistic encoder, 
$(\boldsymbol{\mu}_{\boldsymbol{z}^{(i)}}, \boldsymbol{\sigma}_{\mathbf{z}^{(i)}}) = f_{\theta}(\mathbf{z}|\mathbf{x}^{(i)})$.
Then samples $L$ latent space vectors, $\boldsymbol{z}^{(i,l)} \sim \mathcal{N}(\boldsymbol{\mu}_{\boldsymbol{z}^{(i)}}, \boldsymbol{\sigma}_{\boldsymbol{z}^{(i)}})$.
The parameters of the input distribution are reconstructed using these $L$ samples, $\boldsymbol{\mu}_{\hat{\boldsymbol{x}}^{(i,l)}}, \boldsymbol{\sigma}_{\hat{\mathbf{x}}^{(i, l)}} = g_{\phi}(\mathbf{x}|\mathbf{z}^{(i,l)})$
then the reconstruction probability for test sample $\mathbf{x}^{(i)}$ is given by :

\begin{equation}
	P_\text{recon}(\mathbf{x}^{(i)}) = \frac{1}{L} \sum_{l=1}^L p_{\theta}\bigg(\mathbf{x}^{(i)} | 
	\boldsymbol{\mu}_{\hat{\mathbf{x}}^{(i,l)}}, \boldsymbol{\sigma}_{\hat{\mathbf{x}}^{(i, l)}}\bigg)
\end{equation}

Multiple samples drawn from the latent variable distribution, lets $P_\text{recon}(\mathbf{x}^{(i)})$ 
take into account the variability of the latent variable space, which is one of the essential distinctions 
between the stochastic variational autoencoder and a standard autoencoder, where latent variables are 
defined by deterministic mappings.

\subsection{Generative Adversarial Networks (GANs)}

A GAN \cite{goodfellow2014generative} consists of a generator $G$, usually a decoder, and a discriminator
$D$, usually an binary classifier that assigns a probability of an image being generated (fake), 
or sampled from the training data (real). 
The generator $G$ in fact learns a distribution $p_g$ over data $\mathbf{x}$ via a mapping $G(\mathbf{z})$ of
samples $\mathbf{z}$, 1D vectors of uniformly distributed input noise sampled from latent
space $\mathcal{Z}$, to 2D images in the image space manifold $\mathcal{X}$ , which is populated by
normal examples. In this setting, the network architecture of the generator $G$
is equivalent to a convolutional decoder that utilizes a stack of strided convolutions. 
The discriminator $D$ is a standard CNN that maps a 2D image to a
single scalar value $D(\cdot)$. The discriminator output $D(\cdot)$ can be interpreted as the
probability that the given input to the discriminator $D$ was a real image $\mathbf{x}$ sampled 
from training data $X$ or a generated image using $G(\mathbf{z})$ by the generator $G$. $D$ and $G$ are 
simultaneously optimized through the following two-player minimax game with
value function $V (G, D)$ :

\begin{equation}
	\min_G \max_D V(D,G) = \mathbb{E}_{\mathbf{x} \sim p_\text{data}} [\log D(\mathbf{x})] + 
						   \mathbb{E}_{\mathbf{z} \sim p_\mathbf{z}(\mathbf{z})} [\log (1-D(G(\mathbf{z})))]
\end{equation}

The discriminator is trained to maximize the probability of assigning real training examples 
the ``real'' and samples from $p_g$ the ``fake'' label. The generator $G$
is simultaneously trained to fool $D$ via minimizing $V(G) = \log (1 - D(G(\mathbf{z})))$,
which is equivalent to maximizing $V(G) = D(G(\mathbf{z}))$.
During adversarial training, the generator improves in generating realistic 
images and the discriminator progresses in correctly identifying real and generated images.
GANs are implicit models \cite{Goodfellow-GANS-NIPS-2016}, that sample directly from the distribution 
represented by the model.

\subsection{GANs for anomaly detection in Images}
This section reviews work done by authors \cite{schlegl2017unsupervised} who apply a GAN model for the
task of anomaly detection in medical images. GANs are generative models that best produce a set of 
training data points $\mathbf{x} \sim P_\text{data}(\mathbf{x})$ 
where $P_\text{data}$ represents the probability density of the training data points. 
The basic idea in anomaly detection is to be able to evaluate the density function 
of the normal vectors in the training set containing no anomalies while for
the test set we evaluate a negative log-likelihood score which serves as the final anomaly 
score. The score corresponds to the test sample's posterior probability of being generated from the same generative model
representing the training data points. GANs provide a generative model that minimizes the distance
between the training data distribution and the generative model samples without explicitly defining 
a parametric function, which is why it is called an implicit generative model \cite{Goodfellow-GANS-NIPS-2016}.
Thus to be successfully used in an anomaly detection framework the authors \cite{schlegl2017unsupervised} 
evaluate the mapping $\mathbf{x} \to \mathbf{z}$, i.e. Image domain $\to$ latent representation.
This was done by choosing the closest point $\mathbf{z}_\gamma$ using back-propagation.
Once done the residual loss in the image space was defined as 
$\mathcal{L}_R (\mathbf{z}_\gamma) = \sum |\mathbf{x}-G(\mathbf{z}_\gamma)|$.

GANs are generative models and to evaluate a likelihood one requires a mapping from the image domain to
the latent space. This is achieved by authors in \cite{schlegl2017unsupervised}, which we shall shortly describe here.
Given a query image $\mathbf{x} \sim p_\text{test}$, the authors aim to find a point $\mathbf{z}$ 
in the latent space that corresponds to an image $G(z)$ that is visually most similar to the query 
image $\mathbf{x}$ and that is located on the manifold $\mathcal{X}$. The degree of similarity 
of $\mathbf{x}$ and $G(\mathbf{z})$ depends on to which extent the query image follows the 
data distribution $p_g$ that was used for training of the generator.

To find the best $\mathbf{z}$, one starts randomly sampling $\mathbf{z}_1$ from the latent 
space distribution $\mathcal{Z}$ and feeds it into the trained generator
which yields the generated image $G(\mathbf{z}_1)$. Based on the generated image $G(\mathbf{z}_1)$ we 
can define a loss function, which provides gradients for the update of the coefficients of $\mathbf{z}_1$
resulting in an updated position in the latent space, $\mathbf{z}_2$ . In order to find the most
similar image $G(\mathbf{z}_\Gamma)$, the location of $\mathbf{z}$ in the latent space $\mathcal{Z}$ 
is optimized in an iterative process via $\gamma = 1, 2, . . . , \Gamma$ back-propagation steps.

\subsection{Adversarial Discriminators using Cross-channel prediction}

Here we shall review the work done in \cite{AdversarialDiscriminators_2017} 
applied to anomaly detection in videos.
The anomaly detection problem in this paper is formulated as a cross-channel
prediction task, where the two channels are the raw-image values $F_t$ and the optical 
flow vectors $O_t$ for frames $F_t, F_{t-1}$ in the videos. This work combines 
two architectures, the pixel-GAN architecture by \cite{pix2pix2016}
to model the normal/training data distribution, and the Split-Brain Autoencoders \cite{zhang2017split}.  
The Split-Brain architectures aims at predicting a multi-channel output by building 
cross-channel autoencoders. That is, given training examples $X \in \mathbb{R}^{H \times W \times C}$, 
we split data into $X_1 \in \mathbb{R}^{H \times W \times C_1}$ and $X_2 \in \mathbb{R}^{H \times W \times C_2}$,
where $C_1, C_2 \subset C$, and the authors train multiple deep representations $\widetilde{X_2} = \mathcal{F}_1(X_1)$
and $\widetilde{X_1} = \mathcal{F}_2(X_2)$, which when concatenated provided a reconstruction of the input tensor $X$,
just like an autoencoder. Various manners of aggregating these predictors have been explored in \cite{zhang2017split}.
In the same spirit as the cross-channel autoencoders \cite{zhang2017split}, Conditional GANs were developed \cite{pix2pix2016}
to learn a generative model that learns a mapping from one input domain to the other.

The authors \cite{AdversarialDiscriminators_2017} train two networks much in the spirit of the conditional GAN \cite{pix2pix2016}
where : $\mathcal{N}^{O \to F}$ which generates the raw image frames from the optical flow and $\mathcal{N}^{F \to O}$ which generates 
the optical flow from  the raw images. $F_t$ are image frames with RGB channels and $O_t$ are vertical 
and horizontal optical flow vector arrays. The input to discriminator $D$ is thus a 6-D 
tensor. We now describe the adaption of the cross-channel autoencoders for the task of anomaly detection.

\begin{itemize}
	\item $\mathcal{N}^{F \to O}$ : Training set is $X = \{(F_t, O_t)\}_{t=1}^N$.
	The L1 loss function with $x = F_t, y = O_t$ :
	\begin{equation}
		\mathcal{L}_{L1}(x, y) = \| y - G(x,z) \|_1
	\end{equation}
	with the conditional adversarial loss being
	\begin{equation}
    	\begin{split}
    		\mathcal{L}_{cGAN}(G, D) = 
            & \mathbb{E}_{(x,y) \in X} [\log D(x,y)] \\ + 
            & \mathbb{E}_{x \in \{F_t\}, z \in \mathcal{Z}} \log (1-D(x, G(x, z)))
    	\end{split}  
	\end{equation}
	\item Conversely in $\mathcal{N}^{O \to F}$ the training set changes to $X = \{(O_t, F_t)\}_{t=1}^N$.
\end{itemize}

The generators/discriminators follow a U-net architecture as in \cite{pix2pix2016} with skip connections.
The two generators $G^{F \to O}, G^{O \to F}$ are trained to map training frames and their optical 
flow to their cross-channel counterparts. The goal is to force a poor cross-channel prediction
on test video frames containing an anomaly so that the trained discriminators shall provide a
low probability score.

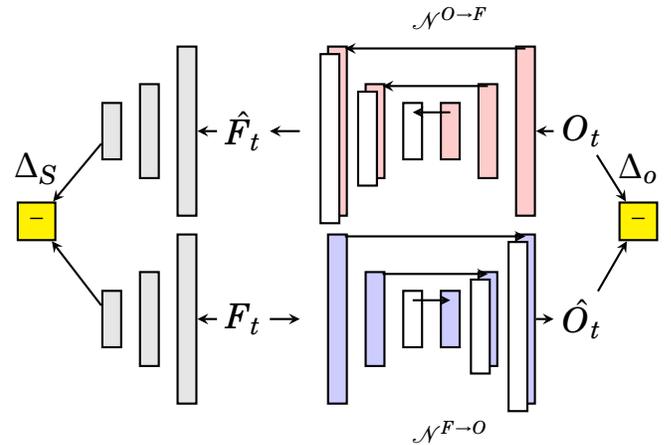
\begin{figure}[htbp]
	\centering
	\begin{tikzpicture}
	\node[draw,thick,minimum width=.25cm,minimum height=2.25cm, fill=red!20]  at (0,0+2.5)  {};
	\node[draw,thick,minimum width=.25cm,minimum height=2.25cm, fill=white]  at (-0.1,-0.1+2.5)  {};
	\node[draw,thick,minimum width=.25cm,minimum height=1.25cm, fill=red!20]  at (0.5,0+2.5)  {};
	\node[draw,thick,minimum width=.25cm,minimum height=1.25cm, fill=white]  at (0.4,-0.1+2.5)  {};
	\node[draw,thick,minimum width=.25cm,minimum height=0.75cm, fill=white]  at (1,0+2.5)  {};
	\node[draw,thick,minimum width=.25cm,minimum height=0.75cm, fill=red!20]  at (1.5,0+2.5)  {};
	\node[draw,thick,minimum width=.25cm,minimum height=1.25cm, fill=red!20]  at (2,0+2.5)  {};
	\node[draw,thick,minimum width=.25cm,minimum height=2.25cm, fill=red!20] (nb)  at (2.5,0+2.5)  {};

	\draw[arrow] (2.5, 1.1+2.5)--(0.1,1.1+2.5);
	\draw[arrow] (2, 0.6+2.5)--(0.6,0.6+2.5);
	\draw[arrow] (1.5, 0.25+2.5)--(1,0.25+2.5);

	\node[draw,thick,minimum width=.25cm,minimum height=2.25cm, fill=blue!20]  at (0,0)  {};
	\node[draw,thick,minimum width=.25cm,minimum height=1.25cm, fill=blue!20]  at (0.5,0)  {};
	\node[draw,thick,minimum width=.25cm,minimum height=0.75cm, fill=white]  at (1,0)  {};
	\node[draw,thick,minimum width=.25cm,minimum height=0.75cm, fill=blue!20]  at (1.5,0)  {};
	\node[draw,thick,minimum width=.25cm,minimum height=1.25cm, fill=blue!20]  at (2,0)  {};
	\node[draw,thick,minimum width=.25cm,minimum height=1.25cm, fill=white]  at (1.9,-0.1)  {};
	\node[draw,thick,minimum width=.25cm,minimum height=2.25cm, fill=blue!20] (na)  at (2.5,0)  {};
	\node[draw,thick,minimum width=.25cm,minimum height=2.25cm, fill=white]  at (2.4,-0.1)  {};

	\draw[arrow] (0.1,1.1)--(2.5, 1.1);
	\draw[arrow] (0.6,0.6)--(2, 0.6);
	\draw[arrow] (1,0.25)--(1.5, 0.25);

	\node[draw,thick,minimum width=.25cm,minimum height=0.75cm, fill=gray!20] (a)  at (-3,0+2.5)  {};
	\node[draw,thick,minimum width=.25cm,minimum height=1.25cm, fill=gray!20] at (-2.5,0+2.5)  {};
	\node[draw,thick,minimum width=.25cm,minimum height=2.25cm, fill=gray!20] (aa) at (-2,0+2.5)  {};

	\node[draw,thick,minimum width=.25cm,minimum height=0.75cm, fill=gray!20] (b) at (-3,0)  {};
	\node[draw,thick,minimum width=.25cm,minimum height=1.25cm, fill=gray!20]  at (-2.5,0)  {};
	\node[draw,thick,minimum width=.25cm,minimum height=2.25cm, fill=gray!20] (bb) at (-2,0)  {};

	\node[draw, thick,minimum width=.5cm,minimum height=0.5cm, fill=yellow]  (c) at (-4,1.3)  {$-$};
	\node[draw, thick,minimum width=.5cm,minimum height=0.5cm, fill=yellow]  (d) at (4,1.3)  {$-$};

	\node at (1.5,4) {$\mathcal{N}^{O\to F}$};
	\node at (1.5,-1.5) {$\mathcal{N}^{F\to O}$};

	\node (e) at (3.25,2.5) {\Large $O_t$};
	\node (f) at (3.25,0) {\Large $\hat{O_t}$};
	\node (g) at (-1.25,0) {\Large $F_t$};
	\node (h) at (-1.25,2.5) {\Large $\hat{F_t}$};

	\node at (-4,2) {\Large $\Delta_S$};
	\node at (4,2) {\Large $\Delta_o$};
	\draw[arrow] (a)--(c);
	\draw[arrow] (b)--(c);
	\draw[arrow] (e)--(d);
	\draw[arrow] (f)--(d);
	\draw[arrow] (g)--(-0.5, 0);
	\draw[arrow] (h)--(aa);
	\draw[arrow] (g)--(bb);
	\draw[arrow] (-0.5, 2.5)--(h);
	\draw[arrow] (na)--(f);
	\draw[arrow] (e)--(nb);
	
	\end{tikzpicture}
	\caption{The cross-channel prediction conditional GAN architecture in 
	\cite{AdversarialDiscriminators_2017}. There are two GAN models : 
	flow$\to$RGB predictor ($\mathcal{N}^{O\to F}$) and RGB$\to$Flow 
	predictor ($\mathcal{N}^{F\to O}$). Each of the generators shown 
	has a U-net architecture which uses the common underlying structure 
	in the image RGB channels and optical flow between two frames.}
	\label{fig:CrossGAN}
\end{figure}

The trained discriminators $D^{F \to O}, D^{O \to F}$ are patch-discriminators
that produce scores $S^O, S^F$ on a grid with resolution smaller than the image.
These scores do not require the reconstruction of the different channels to be evaluated.
The final score is $S = S^O + S^F$ which is normalized between $[0,1]$ based on the maximum 
value of individual scores for each frame.
The U-net uses the Markovian structure present spatially by the skip connections shown between 
the input and the output of the generators in figure \ref{fig:CrossGAN}.
Cross-channel prediction aims at modeling the spatio-temporal correlation present across channels 
in the context of video anomaly detection.

\subsection{Adversarial Autoencoders (AAEs)}  
Adversarial Autoencoders are probabilistic autoencoders that use GANs to perform
variational approximation of the aggregated posterior of the latent 
space representation \cite{Alireza_AAE_2016_ICLR} using an arbitrary prior. 

\begin{figure}[htbp]
	\centering
	\begin{tikzpicture}[node distance=1cm]
	\node[draw,thick,minimum width=1cm,minimum height=1cm] (x) at (1,0)  {$\mathbf{x}$};
	\node[draw,thick,minimum width=0.5cm,minimum height=2cm] (h1) at (2,0)  {};
	\node[draw,thick,minimum width=0.5cm,minimum height=2cm] (h2) at (3,0)  {};
	\node[draw,thick,minimum width=0.5cm,minimum height=1cm, fill=gray] (h) at (4,0)  {};
	\node[draw,thick,minimum width=0.5cm,minimum height=2cm] (h2d) at (5,0)  {};
	\node[draw,thick,minimum width=0.5cm,minimum height=2cm] (h1d) at (6,0)  {};
	\node[draw,thick,minimum width=1cm,minimum height=1cm] (xrec) at (7,0)  {$\widehat{\mathbf{x}}$};
	\node[draw,thick,minimum width=1cm,minimum height=0.5cm] (input) at (4,-3) {Input};
	\node[draw,thick,minimum width=1cm,minimum height=0.5cm, fill=yellow] (pz) at (2,-3) {$p(\mathbf{z})$};
	\node[draw,thick,minimum width=0.5cm,minimum height=2cm] (disch2d) at (5,-3) {};
	\node[draw,thick,minimum width=0.5cm,minimum height=2cm] (disch1d) at (6,-3) {};
	\node[draw,thick,minimum width=0.5cm,minimum height=0.5cm] (bottomout) at (7,-3) {$\frac{1}{1+e^{-x}}$};

	\node[draw,fill=yellow] at (2.5,1.5) {$q(\mathbf{z}|\mathbf{x})$};
	\node at (4,1) {$\mathbf{z} \sim q(\mathbf{z})$};


	\draw [arrow] (x) -- (h1);
	\draw [arrow] (h1) -- (h2);
	\draw [arrow] (h2) -- (h);
	\draw [arrow] (h) -- (h2d);
	\draw [arrow] (h2d) -- (h1d);
	\draw [arrow] (h1d) -- (xrec);
	\draw [arrow] (h) -- (input) node[midway,left]{$-$};
	\draw [arrow] (pz) -- (input) node[midway,above]{$+$};
	\draw [arrow] (input) -- (disch2d);
	\draw [arrow] (disch2d) -- (disch1d);
	\draw [arrow] (disch1d) -- (bottomout);
	\end{tikzpicture}	
	\caption{Two paths in the Adversarial Autoencoder : Top path refers to the standard 
	autoencoder configuration that minimizes reconstruction error.
	The bottom path constitutes of an adversarial network that ensures an 
	approximation of the user input defined samples from distribution $p(\mathbf{z})$,
	and the latent or code vector distribution, provided by $q(\mathbf{z}|\mathbf{x})$. }
	\label{fig:AAE}
\end{figure}
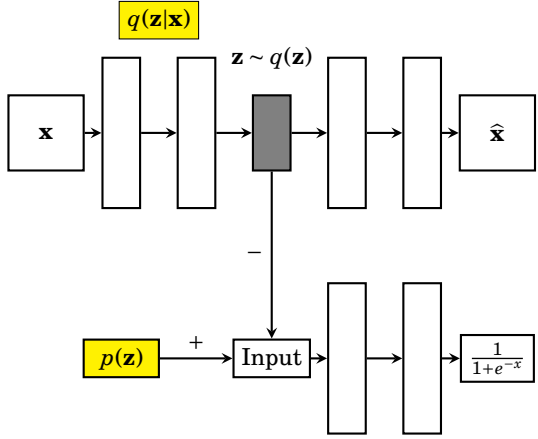

AAEs were applied to the problem of anomalous event detection over images by authors \cite{dimokranitou2017adversarial}.
In figure \ref{fig:AAE}, $\mathbf{x}$ denotes input vectors from training distribution,
$q(\mathbf{z}|\mathbf{x})$ the encoder's posterior distribution, $p(\mathbf{z})$ the prior
that the user wants to impose on the latent space vectors $\mathbf{z}$. The latent space distribution is given by
\begin{equation}
	q(\mathbf{z}) = \int_{\mathbf{x} \in X_\text{train}} q(\mathbf{z}|\mathbf{x}) p_d(\mathbf{x}) d\mathbf{x}
\end{equation}
where $p_d(\mathbf{x})$ represents the training data distribution.
In an AAE, the encoder acts like the generator of the adversarial network, 
and it tries to fool the discriminator into believing that $q(\mathbf{z})$
comes from the actual data distribution $p(\mathbf{z})$.
During the joint training, the encoder is updated to improve the reconstruction error 
in the autoencoder path, while it is updated by the discriminator of the adversarial network
to make the latent space distribution approach the imposed prior.
As prior distribution for the generator of the network, authors \cite{dimokranitou2017adversarial} use the Gaussian distribution
of 256 dimensions, with the dropout set to 0.5 probability. The method achieves close to state of the art performance.
As the authors remark themselves, the AAE does not take into account the temporal structure in the video sequences. 

\subsection{Controlling reconstruction for anomaly detection}

One of the common problems using deep autoencoders is their capability to produce
low reconstruction errors for test samples, even over anomalous events. This is due
to the way autoencoders are trained in a semi-supervised way on videos with no anomalies, 
but with sufficient training samples, they are able to approximate most test samples well.

In \cite{munawar2017limitingGANs}, the authors propose to limit the reconstruction capability of the generative 
adversarial networks by learning conflicting objectives for the normal and anomalous data.
They use negative examples to enforce explicit poor reconstruction. Thus this setup is weakly supervised, 
not requiring labels.
Given two random variables $X, Y$ with samples $\{\mathbf{x}\}_{i=1}^K, \{\mathbf{y}\}_{j=1}^J$, 
we want the network to reconstruct the input distribution $X$ while poorly reconstruct $Y$.
This was achieved by maximizing the following objective function:

\begin{equation}
	p_{\theta}(\hat{X}|X)-p_{\theta}(\hat{Y}|Y)=
	\sum_{i=1}^K \log p_{\theta}(\hat{\mathbf{x}_i}|\mathbf{x}_i)+
	\sum_{j=1}^J \log p_{\theta}(\hat{\mathbf{y}_i}|\mathbf{y}_j)
\end{equation}

where $\theta$ refers to the autoencoders parameters.
This setup assumes strong class imbalance, i.e.very few samples of the anomalous class $Y$ 
are available compared to the normal class $X$.
The motivation for negative learning using anomalous examples
is to consistently provide poor reconstruction of anomalous samples. 
During the training phase, authors \cite{munawar2017spatiotemporalAnomaly} 
reconstruct positive samples by minimizing the reconstruction error 
between samples, while negative samples are forced to have a bad 
reconstruction by maximizing the error. This last step was termed as
negative learning. The datasets evaluated were the reconstruction 
of the images from MNIST and Japanese highway video patches \cite{japanhighway2017youtube}.

In similar work by \cite{razakarivony2014discriminative}, discriminative autoencoders aim
at learning low-dimensional discriminative representations for positive ($X^+$) and negative ($X^-$)
classes of data. The discriminative autoencoders build a latent space representation 
under the constraint that the positive data should be better reconstructed than the negative data.
This is done by minimizing the reconstruction error for positive examples while ensuring 
that those of the negative class are pushed away from the manifold.

\begin{equation}
	\mathcal{L}_d(X^+ \cup X^-) = \sum_{\mathbf{x} \in X^+ \cup X^-} \max(0, t(\mathbf{x})\cdot (\|\hat{\mathbf{x}}-\mathbf{x}\|-1))
	\label{eqn:discae}
\end{equation}
In the above loss function, $t(\mathbf{x}) = \{-1,+1\}$ denotes as the label of the sample, and 
$e(\mathbf{x}) = \|\mathbf{x}-\widetilde{\mathbf{x}}\|$ the distance of that example
to the manifold. Minimizing the hinge loss in equation \ref{eqn:discae} achieves reconstruction 
such that the discriminative autoencoders build a latent space representation of
data that better reconstructs positive data compared to the negative data

\section{Experiments}
There are two large classes of experiments : first, reconstructing the input video 
on a single frame basis $X_t \to \widetilde{X}_t$, second the reconstruction of a 
stack of frames $X_{t-p:t} \to \widetilde{X}_{t-p:t}$. These reconstruction schemes
are performed either on raw frame values, or on the optical flow between consequent frame pairs.
Reconstructing raw image values modeled the back-ground image, since minimizing the reconstruction
error was in fact evaluating the background. Convolutional autoencoders reconstructing a sequence
of frames captured  temporal appearance changes as described by \cite{hasan2016learning}.
When learning feature representations on optical flow we indirectly operate on two frames, 
since each optical flow map evaluates the relative motion between two consequent frame pairs. 
In the case of predictive models the current frame $X_t$ was predicted after observing the past $p$
frames. This provides a different temporal structure as compared to a simple reconstruction of a sequence
of frames $X_{t-p:t} \to \widetilde{X}_{t-p:t}$, where the temporal coherence results from enforcing 
a bottleneck in the autoencoder architectures. The goal of these experiments were not evaluate the 
best performing model, and were intended as a tool to understand how background estimation 
and temporal appearance were approximated by the different models. A complete detailed study is beyond 
the scope of this review.

In this section, we evaluate the performance of the following classes of models on the UCSD and CUHK-Avenue datasets. 
As a baseline, we use the reconstruction of a dense optical flow calculated using the Farneback method in OpenCV 3,
by principal component analysis, with around 150 components. For predictive models, as a baseline we use 
a vector autoregressive model (VAR), referred to as \textit{LinPred}. The coefficients of the model are 
estimated on a lower dimensional, random projection of the raw image or optical flow maps from the 
input training video stream. The random projection avoids badly conditioned and expensive matrix inversion. 
We compare the performance of Contractive autoencoders, simple 3D autoencoders 
based on C3D \cite{tran2015learning_C3D} CNNs (C3D-AE), the ConvLSTM and ConvLSTM autoencoder from the 
predictive model family and finally the VAE from the generative models family. The VAE's loss function
consists of the binary cross-entropy (similar to a reconstruction error) between the model prediction 
and the input image, and the KL-divergence $D_{KL}[Q(z|X)\|P(z)]$, between the encoded latent space 
vectors and $P(z) = \mathcal{N}(0,I)$ multivariate unit-Gaussian.
These models were built in Keras \cite{chollet2015keras} with Tensorflow back-end and executed on a K-80 GPU.

\subsection{Architectures}
Our Contractive and Variational AE (VAE) constitutes of a random projection to reduce the 
dimensionality to 2500 from an input frame of size 200 $\times$ 200.
The Contractive AE constitutes of one fully connected hidden layer of 
size $1250$ which map back to the reconstruction of the randomly projected vector of size 2500.
While the VAE contains two hidden layers (dimensions: 1024, 32), 
which maps back to the output of 2500 dimensions. We use the latent space representation 
of the variational autoencoders to fit a multivariate 1-Gaussian on the training dataset 
and evaluate the negative-log probability for the test samples.

\begin{table*}
\caption{Area Under-ROC (AUROC)}
\label{tab:auroc}
\begin{center}
	\begin{tabular}{@{}*{8}{c}@{}}
	\hline
	 \thead{Methods\\ Feature} & \thead{PCA\\ flow} & \thead{LinPred \\ (raw, flow)} & \thead{C3D-AE \\ (raw, flow)} & \thead{ConvLSTM \\ (raw, flow)} & \thead{ConvLSTM-AE \\ (raw, flow)} & \thead{CtractAE \\ (raw, flow)} & \thead{VAE \\ (raw, flow)} \\ 
	\hline
	UCSDped1    & 0.75 & (0.71, 0.71) & (0.70, 0.70) & (0.67, 0.71) & (0.43, 0.74) & (0.66, \textbf{0.75}) & (0.63, 0.72) \\ 
	UCSDped2    & 0.80 & (0.73, 0.78) & (0.64, 0.81) & (0.77, 0.80) & (0.25, 0.81) & (0.65, 0.79) & (0.72, \textbf{0.86}) \\ 
	CUHK-avenue & 0.82 & (0.74, 0.84) & (\textbf{0.86}, 0.18) & (0.84, 0.18) & (0.50, 0.84) & (0.83, 0.84) & (0.78, 0.80) \\ 
	 \hline
 	\end{tabular}
\end{center}
\end{table*}

\begin{table*}
\caption{Area Under-Precision Recall (AUPR)}
\label{tab:aupr}
\begin{center}
	\begin{tabular}{@{}*{8}{c}@{}}
	\hline
	 \thead{Methods\\ Feature} & \thead{PCA\\ flow} & \thead{LinPred \\ (raw, flow)} & \thead{C3D-AE \\ (raw, flow)} & \thead{ConvLSTM \\ (raw, flow)} & \thead{ConvLSTM-AE \\ (raw, flow)} & \thead{CtractAE \\ (raw, flow)} & \thead{VAE \\ (raw, flow)} \\ 
	\hline
	UCSDped1    & 0.78 & (0.71, 0.71) & (0.75, 0.77) & (0.67, 0.71) & (0.52, 0.81) & (0.70, \textbf{0.81}) & (0.66, 0.76) \\ 
	UCSDped2    & 0.95 & (0.92, 0.94) & (0.88, 0.95) & (0.94, 0.95) & (0.74, 0.95) & (0.88, 0.95) & (0.92, \textbf{0.97}) \\ 
	CUHK-avenue & 0.66 & (0.49, 0.65) & (0.65, 0.15) & (0.66, 0.15) & (0.34, 0.70) & (0.69, \textbf{0.70}) & (0.54, 0.68) \\ 
	 \hline
 	\end{tabular}
\end{center}
\end{table*}

\subsection{Observations and Issues}

The results are summarized in the table \ref{tab:auroc} and \ref{tab:aupr}. 
The performance measures reported are the Area Under 
Receiver-Output-Characteristics plot (AU-ROC), Area Under Precision-Recall plot. 
These scores are calculated when the input channels correspond to 
the raw image (raw) and the optical flow (flow), each of which has 
been normalized by the maximum value. The final temporal anomaly score is given 
by equation \ref{eqn:anomaly_score}. These measures are described in 
the next section. We also describe the utility of these measures 
under different frequencies of occurrences of the anomalous positive class.

\textbf{Reconstruction model issues :} 
Deep autoencoders identify anomalies by poor reconstruction of objects that have 
never appeared in the training set, when raw image pixels are used as input.
It is difficult to achieve this in practice due to a stable reconstruction of 
new objects by deep autoencoders. This pertains to the high capacity of autoencoders, 
and their tendency to well approximate even the anomalous objects. Controlling 
reconstruction using negative examples could be a possible solution. This holds true, 
but to a lower extent, when reconstructing a sequence of frames (spatio-temporal block).

\textbf{AUC-ROC vs AUC-PR :}
The anomalies in the UCSD pedestrian dataset have a duration of several hundred 
frames on average, compared to the anomalies in the CUHK avenue dataset
which occur only for a few tens of frames. This makes the anomalies statistically 
less probable. This can be seen by looking at the AU-PR table \ref{tab:aupr},
where the average scores for CUHK-avenue are much lower than for UCSD pedestrian 
datasets. It is important to note that this does not mean the performance
over the CUHK-Avenue dataset is lower, but just the fact that the positive 
anomalous class is rarer in occurrence.	

\textbf{Rescaling image size :} 
The models used across different experiments in the articles that were reviewed, 
varied in the input image size. In some cases the images were resized
to sizes (128,128), (224, 224), (227, 227). We have tried to fix this to be 
uniformly (200,200). Though it is essential to note that there is 
a substantial change in performance when this image is resized to certain sizes.

\textbf{Generating augmented video clips :} 
Training the convolutional LSTM for video anomaly detection takes a large number of epochs. 
Furthermore, the training video data required is much higher and data augmentation
for video anomaly detection requires careful thinking. Translations and rotations 
may be transformations to which the anomaly detection algorithm requires to be sensitive to, 
based on the surveillance application. 

\textbf{Performance of models :} VAEs perform consistently as well as or better than PCA on optical flow.
It is still left as a future study to understand clearly, why the performance of a stochastic autoencoder such as VAE is better.
Convolutional LSTM on raw image values follow closely behind as the first predictive model performing as 
good as PCA but sometimes poorer. Convolutional LSTM-AE is a similar architecture with similar performance. 
Finally, the 3D convolutional autoencoder, based on the work by \cite{tran2015learning_C3D}, performs as 
well as PCA on optical flow, while modeling local motion patterns.

To evaluate the specific advantages of each of these models, a larger number of real world, 
video surveillance examples are required demonstrating the representation or feature that is most discriminant.
In experiments, we have also observed that application of PCA on the random projection of 
individual frames performed well in avenue dataset, indicating that very few frames were 
sufficient to identify the anomaly; while the PCA performed poorly on 
UCSD pedestrian datasets, where motion patterns were key to detect the anomalies.

For single frame input based models, optical flow served as a good input since it already encoded part of 
the predictive information in the training videos. On the other hand, convolutional LSTMs and 
linear predictive models required $p=[2,10]$ input raw image values, in the training videos, 
to predict the current frame raw image values.

\subsection{Evaluation measures}

The anomaly detection task is a single class estimation task, where   
$0$ is assigned to samples with likelihood (or reconstruction error) above (below) a certain threshold, 
and $1$ assigned to detected anomalies with low likelihood (high reconstruction error).
Statistically, the anomalies are a rare class and occurs less frequently
compared to the normal class. The most common characterization of this behavior
is the expected frequency of occurrence of anomalies.
We briefly review the anomaly detection evaluation procedure as well as the performance 
measures that were used across different studies. 
For a complete treatment of the subject, the reader is referred to \cite{fawcett2006introduction}.

The final anomaly score is a value that is treated as a probability which lies in $s(t) \in [0,1], \forall t$, 
$t \in [1,T]$, $T$ being the maximum time index.
For various level sets or thresholds of the anomaly score $a_{1:T}$, one can evaluate the 
True Positives (TP, the samples which are truly anomalous and detected as anomalous), 
True Negatives (TN, the samples that are truly normal and detected as normal), 
False Positives (FP, the samples which are truly normal samples but detected as anomalous)
and finally False Negatives (FN, the samples which are truly anomalous but detected as normal).
\begin{equation}
\begin{split}
	\text{True positive rate (TPR) or Recall} = & \frac{\sum \text{TP}}{\text{TP+FN}}\\
	\text{False positive rate(FPR)} = & \frac{\sum \text{FP}}{\text{TP+FN}}\\
	\text{Precision} = & \frac{\sum \text{TP}}{\text{TP+FP}}
\end{split}
\end{equation}
We require these measures to evaluate the ROC curve, which measures the performance 
of the detection at various False positive rates. That is ROC plots TPR vs FPR, while the PR
plots precision vs recall.

The performance of anomaly detection task is evaluated based on an important criterion, 
the probability of occurrence of the anomalous positive class. Based on this value, different
performance curves are useful. We define the two commonly used performance curves : 
\textit{Precision-Recall} (PR) curves and \textit{Receiver-Operator-Characteristics} (ROC) curves.
The area under the PR curve (AU-PR) is useful when true 
negatives are much more common than true positives (i.e., TN >> TP).
The precision recall curve only focuses on predictions around the positive (rare) class. 
This is good for anomaly detection because predicting true negatives (TN) is easy in anomaly detection. 
The difficulty is in predicting the rare true positive events.
Precision is directly influenced by class (im)balance since FP is affected, whereas TPR only 
depends on positives. This is why ROC curves do not capture such effects. Precision-recall 
curves are better to highlight differences between models for highly imbalanced data sets. 
For this reason, if one would like to evaluate models under imbalanced class settings, AU-PR scores
would exhibit larger differences than the area under the ROC curve.

\section{Conclusion}

In this review paper, we have focused on categorizing the different unsupervised learning models 
for the task of anomaly detection in videos into three classes based on the prior information used to build the 
representations to characterize anomalies.

They are reconstruction based, spatio-temporal predictive models, and generative models.
Reconstruction based models build representations that minimize the reconstruction error 
of training samples from the normal distribution. Spatio-temporal predictive models take 
into account the spatio-temporal correlation by viewing videos as a spatio-temporal time series.
Such models are trained to minimize the prediction error on spatio-temporal sequences from the training series, 
where the length of the time window is a parameter. Finally, the generative models learn to generate
samples from the training distribution, while minimizing the reconstruction error as well as distance between 
generated and training distribution, where the focus is on modeling the distance between sample and distributions. 

Each of these methods focuses on learning certain prior information that is useful for constructing the representation 
for the video anomaly detection task.  One key concept which occurs in various architectures for video anomaly detection 
is how temporal coherence is implemented. Spatio-temporal autoencoders and Convolutional LSTM learn a reconstruction based or 
spatio-temporal predictive model that both use some form of (not explicitly defined) spatio-temporal regularity assumptions.
We can conclude from our study that evaluating how sensitive the learned representation is to certain 
transformations such as time warping, viewpoint, applied to the input training video stream, 
is an important modeling criterion. Certain invariances are as well defined by the choice of the 
representation (translation, rotation) either due to reusing convolutional architectures 
or imposing a predictive structure. A final component in the design of the video anomaly 
detection system is the choice of thresholds for the anomaly score, which was not covered 
in this review. The performance of the detection systems were evaluated using ROC plots which 
evaluated performance across all thresholds. Defining a spatially variant threshold is an 
important but non-trivial problem.

Finally, as more data is acquired and annotated in a video-surveillance 
setup, the assumption of having no labeled anomalies progressively turns false, partly discussed 
in the section on controlling reconstruction for anomaly detection. Certain anomalous points 
with well defined spatio-temporal regularities become a second class that can be estimated well; 
and methods to include the positive anomalous class information into detection 
algorithms becomes essential. Handling class imbalance becomes essential in such a case.

Another problem of interest in videos is the variation in temporal scale of motion patterns 
across different surveillance videos, sharing a similar background and foreground. Learning 
a representation that is invariant to such time warping would be of practical interest.

There are various additional components of the stochastic gradient descent algorithm that were not covered in this review.
The Batch Normalization \cite{BatchNormIoffe15} and drop-out based regularization \cite{srivastava2014dropout}
play an important role in the regularization of deep learning architectures, and a systematic study is important
to be successful in using them for video anomaly detection.

\textbf{Acknowledgments :}
The authors would like to thank Benjamin Crouzier for his help in proof reading the manuscript, 
and Y. Senthil Kumar (Valeo) for helpful suggestions. The authors would also like to thank their employer
to perform fundamental research.



\bibliographystyle{IEEEtran}
\bibliography{bibliography}
\end{document}